\documentclass[10pt,twocolumn,letterpaper]{article}

\usepackage[pagenumbers]{cvpr} % To force page numbers, e.g. for an arXiv version

\usepackage{graphicx}
\usepackage{amsmath}
\usepackage{amssymb}
\usepackage{booktabs}
\usepackage{times}
\usepackage{epsfig}
\usepackage{bm}
\usepackage{multirow}
\usepackage{cuted}
\usepackage{capt-of}
\usepackage{caption}
\usepackage{animate}
\usepackage{bbding}
\usepackage{footnote}
\usepackage{xcolor}
\usepackage{diagbox}
\usepackage{colortbl}

\makeatletter
\renewcommand\@makefnmark{\hbox{\@textsuperscript{\normalfont\color{red}\@thefnmark}}}
\makeatother

\usepackage[pagebackref,breaklinks,colorlinks]{hyperref}

\usepackage[capitalize]{cleveref}
\crefname{section}{Sec.}{Secs.}
\Crefname{section}{Section}{Sections}
\Crefname{table}{Table}{Tables}
\crefname{table}{Tab.}{Tabs.}

\graphicspath{{images/}}

\newcommand{\paravspace}{\vspace{-12pt}}

\definecolor{yellow}{rgb}{1, 1, 0.7}
\definecolor{orange}{rgb}{1, 0.85, 0.7}
\definecolor{pink}{rgb}{1, 0.7, 0.7}

\begin{document}

% \title{PortraitGRAM: One-Shot and 3D-Consistent Portrait Synthesis via Efficient Generative Radiance Manifolds}
\title{Learning Detailed Radiance Manifolds for High-Fidelity and 3D-Consistent Portrait Synthesis from Monocular Image}

\author{Yu Deng \quad Baoyuan Wang \quad Heung-Yeung Shum\\
	Xiaobing.AI
}

\maketitle

% \vspace{-40pt}
\begin{strip}
\vspace{-40pt}
	\centering
	\includegraphics[width=1.0\textwidth]{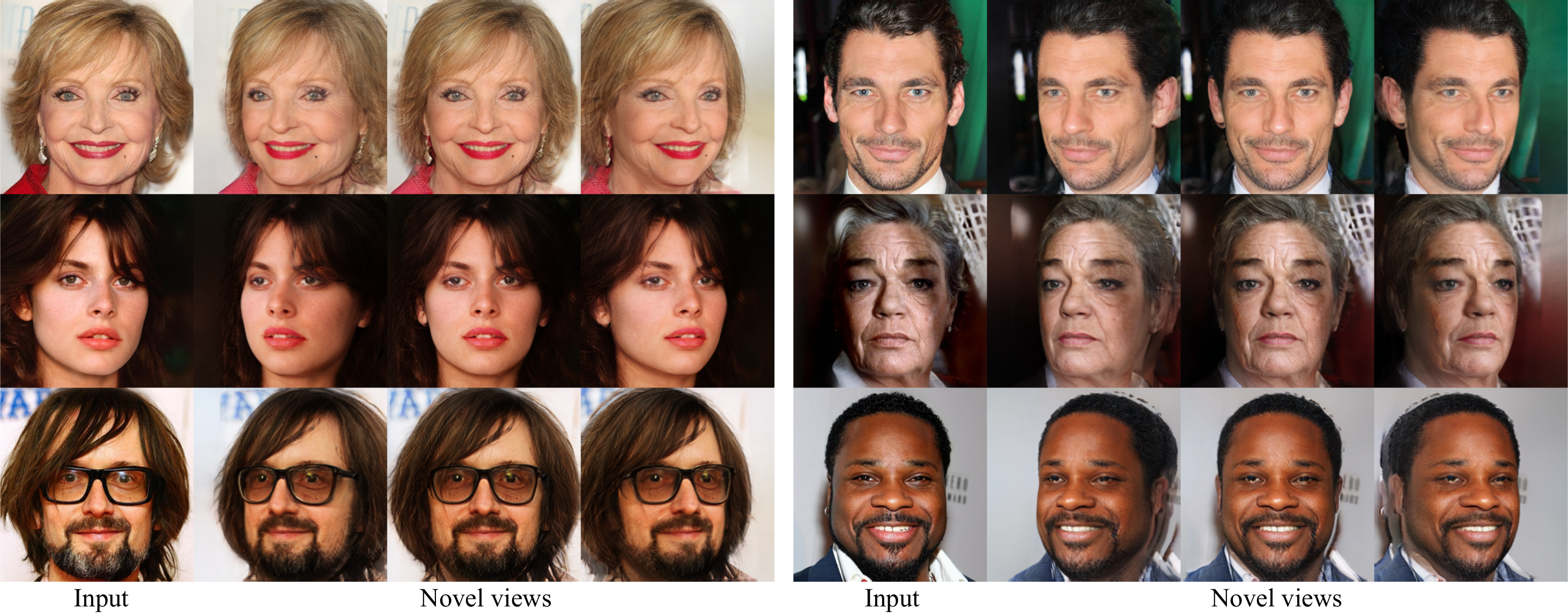}\\
	\captionsetup{type=figure,font=small}
	\vspace{-5pt}
	\caption{Novel view synthesis results by our method. It can generate novel views of a portrait image with high-fidelity and strong 3D-consistency via single forward pass (\eg, see the bangs and wrinkles). \textbf{Best viewed with zoom-in.}
	\label{fig:teaser}
		}
\end{strip}
\vspace{-4pt}

%%%%%%%%% ABSTRACT
\begin{abstract}
\vspace{-8pt}
A key challenge for novel view synthesis of monocular portrait images is 3D consistency under continuous pose variations. Most existing methods rely on 2D generative models which often leads to obvious 3D inconsistency artifacts. We present a 3D-consistent novel view synthesis approach for monocular portrait images based on a recent proposed 3D-aware GAN, namely Generative Radiance Manifolds (GRAM)~\cite{deng2022gram}, which has shown strong 3D consistency at multiview image generation of virtual subjects via the radiance manifolds representation. However, simply learning an encoder to map a real image into the latent space of GRAM can only reconstruct coarse radiance manifolds without faithful fine details, while improving the reconstruction fidelity via instance-specific optimization is time-consuming. We introduce a novel detail manifolds reconstructor to learn 3D-consistent fine details on the radiance manifolds from monocular images, and combine them with the coarse radiance manifolds for high-fidelity reconstruction.
The 3D priors derived from the coarse radiance manifolds are used to regulate the learned details to ensure reasonable synthesized results at novel views. Trained on in-the-wild 2D images, our method achieves high-fidelity and 3D-consistent portrait synthesis largely outperforming the prior art. Project page: \href{https://yudeng.github.io/GRAMInverter}{https://yudeng.github.io/GRAMInverter}

%  A novel manifolds reconstructor is introduced which consists of two branches, one for inverting a given image into the latent space of a pre-trained efficient GRAM to obtain coarse radiance manifolds, the other for reconstructing fine details on the manifolds that cannot be well captured by the coarse result. 3D priors derived from the coarse radiance manifolds are used to regulate the learned details to ensure reasonable synthesized results at novel views. Trained on in-the-wild 2D images, our method achieves high-fidelity and 3D-consistent portrait synthesis largely outperforming prior art.
\end{abstract}

%%%%%%%%% BODY TEXT
\section{Introduction}
\label{sec:intro}

Synthesizing photorealistic portrait images of a person from an arbitrary viewpoint is an important task that can benefit diverse downstream applications such as virtual avatar creation and immersive online communication. Thanks to the thriving of 2D Generative Adversarial Networks (GANs)~\cite{goodfellow2014generative,karras2019style, karras2020analyzing}, people can now generate high-quality portraits at desired views given only monocular images as input, via a simple \textit{invert-then-edit} strategy by conducting GAN inversion~\cite{abdal2019image2stylegan,tov2021designing,roich2021pivotal} and latent space editing~\cite{shen2020interpreting,harkonen2020ganspace, deng2020disentangled,tewari2020pie}. However, existing 2D GAN-based methods 
% This greatly improves the efficiency of portrait synthesis process upon traditional methods which require tedious manual work to design delicate 3D models for physical-based rendering~\cite{cook1982reflectance}.
% Notwithstanding the great success made by 2D GANs for controllable portrait synthesis, 
still have deficiencies when applied to applications that require more strict 3D consistency (\eg VR\&AR). Due to the non-physical rendering process of the 2D CNN-based generators, their synthesized images under pose changes usually bear certain kinds of multiview inconsistency, such as geometry distortions~\cite{abdal2021styleflow,alaluf2022third} and texture sticking or flickering~\cite{karras2021alias,wang2022high}. These artifacts may not be significant enough when inspecting each static image but can be easily captured by human eyes under continuous image variations. 
% Therefore, more advanced methods are still required to achieve 3D-consistent pose control of portrait images.

Recently, there are an emerging group of 3D-aware GANs~\cite{schwarz2020graf,chan2021pi,gu2021stylenerf, chan2021efficient,deng2022gram,schwarz2022voxgraf} targeting at image generation with 3D pose disentanglement. By incorporating Neural Radiance Field (NeRF)~\cite{mildenhall2020nerf} and its variants into the adversarial learning process of GANs, they can produce realistic images with strong 3D-consistency across different views, given only a set of monocular images as training data. As a result, 3D-aware GANs have shown greater potential than 2D GANs for pose manipulations of portraits. However, even though 3D-aware GANs are capable of generating 3D-consistent portraits of virtual subjects, leveraging them for real image pose editing is still a challenging task. To obtain faithful reconstructions of real images, most existing methods~\cite{chan2021pi,deng2022gram,chan2021efficient,sun2022fenerf,sun2022ide,lin20223d,zhang2022training} turn to a time-consuming and instance-specific optimization to invert the given images into the latent space of a pre-trained 3D-aware GAN, which is hard to scale-up. And simply enforcing an encoder-based 3D-aware GAN inversion~\cite{cai2022pix2nerf,sun2022ide} often fails to preserve fine details in the original image.

% an under-explored newborn direction. So far, only a few attempts~\cite{chan2021pi,deng2022gram,chan2021efficient,sun2022ide,cai2022pix2nerf,sun2022fenerf} have tried inversion of a 3D-aware GAN for pose manipulations of a real image, yet they either require carefully tuned optimization process case-by-case~\cite{deng2022gram,chan2021efficient,lin20223d}, or can hardly preserve fine details in the original image~\cite{chan2021pi,cai2022pix2nerf,sun2022fenerf}, or sacrifice certain kinds of 3D-consistency due to leveraging a 2D super-resolution module~\cite{sun2022ide,zhang2022training}. These issues 
% weaken their efficacy in large-scale real scenarios.

In this paper, we propose a novel approach \textit{GRAMInverter}, for high-fidelity and 3D-consistent novel view synthesis of monocular portraits via single forward pass. Our method is built upon the recent GRAM~\cite{deng2022gram} that can synthesize high-quality virtual images with strong 3D consistency via the radiance manifolds representation~\cite{deng2022gram}. Nevertheless, GRAM suffers from the same lack-of-fidelity issue when combined with a general encoder-based GAN inversion approach~\cite{tov2021designing}. The main reason is that the obtained semantically-meaningful low-dimensional latent code cannot well record detail information of the input, as also indicated by some recent 2D GAN inversion methods~\cite{tov2021designing,wang2022high}.

To tackle this problem, our motivation is to further learn 3D-space high-frequency details and combine them with the coarse radiance manifolds obtained from the general encoder-based inversion of GRAM, to achieve faithful reconstruction and 3D-consistent view synthesis.
% This spirit also appears in some recent 2D GAN inversion methods~\cite{wang2022high,yao2022feature}, yet leveraging it for 3D-aware GANs is more challenging as it requires to predict fine details in 3D space from monocular input. 
A straightforward way to achieve this is to extract a high-resolution 3D voxel from the input image and combine it with the coarse radiance manifolds. However, this is prohibited by modern GPUs due to the high memory cost of the 3D voxel. To tackle this problem, we turn to learn a high resolution detail manifolds, taking the advantage of the radiance manifolds representation of GRAM, instead of learning the memory-consuming 3D voxel. We introduce a novel detail manifolds reconstructor to extract detail manifolds from the input images. It leverages manifold super-resolution~\cite{xiang2022gram} to predict high-resolution detail manifolds from a low resolution feature voxel. This can be effectively achieved by a set of memory-efficient 2D convolution blocks. The obtained high resolution detail manifolds can still maintain strict 3D consistency due to lying in the 3D space. We also propose dedicated losses to regulate the detail manifolds via 3D priors derived from the coarse radiance manifolds, to ensure reasonable novel view results.

Another contribution of our method is an improvement upon the memory and time-consuming GRAM, without which it is difficult to be integrated into our GAN inversion framework. We replace the original MLP-based radiance generator~\cite{deng2022gram} in GRAM with a StyleGAN2~\cite{karras2020analyzing}-based tri-plane generator proposed by~\cite{chan2021efficient}. The efficient GRAM requires only $1/4$ memory cost with $7 \times$ speed up, without sacrificing the image generation quality and 3D consistency.

% Based on the efficient GRAM, we design a dedicated GAN inversion scheme for high-fidelity reconstruction and geometrically-consistent novel view synthesis of portrait images. A well-known issue of GAN inversion is a trade-off between fidelity and editability~\cite{tov2021designing}. Inverting images into high-rate latent codes or feature maps gives faithful reconstruction results but can produce severe artifacts after editing, while representing them via low-rate latent codes improves the editing quality at the sacrifice of reconstruction accuracy. To tackle this dilemma, we take inspiration from the latest progress of 2D GAN inversions~\cite{tov2021designing,roich2021pivotal,wang2022high} and design a two-branch image encoder for the efficient GRAM. The first branch is an e4e-like~\cite{tov2021designing} encoder which maps an input image to the low-rate latent codes of the generator to obtain reasonable radiance manifolds for novel pose synthesis; the second branch predicts high-frequency details on the manifolds that cannot be well described by the low-rate latent codes to improve reconstruction fidelity. To ensure that the reconstructed details are reasonable under novel views, we propose multiple losses to regulate them via leveraging 3D priors from our pre-trained efficient GRAM. The whole framework is learned with a multi-stage training scheme given only monocular in-the-wild images as training data.

We train our method on FFHQ dataset~\cite{karras2019style} and conduct multiple experiments to demonstrate its advantages on pose control of portrait images. Once trained, GRAMInverter takes a monocular image as input and predicts its radiance manifolds representation for novel view synthesis at $3$ FPS on a single GPU. The generated novel views well preserve fine details in the original image with strong 3D consistency, outperforming prior art by a large margin. We believe our method takes a solid step towards efficient 3D-aware content creation for real applications.

\section{Related Work}
\paragraph{3D-aware generative model.} Learned with monocular 2D images, 3D-aware GANs~\cite{nguyen2019hologan,schwarz2020graf,niemeyer2021giraffe,chan2021pi,xu2021generative,deng2022gram,chan2021efficient,gu2021stylenerf,or2022stylesdf,xiang2022gram,skorokhodov2022epigraf,schwarz2022voxgraf,gao2022get3d} achieve an explicit disentanglement of camera pose by introducing underlying 3D representations. Earlier works~\cite{nguyen2019hologan,szabo2019unsupervised,shi2021lifting} utilize voxel or mesh as the intermediate representation. Later works~\cite{schwarz2020graf,chan2021pi,xu2021generative,or2022stylesdf,chan2021efficient,deng2022gram,zhao2022generative} leverage NeRF~\cite{mildenhall2020nerf} and its variants~\cite{oechsle2021unisurf,wang2021neus, yariv2021volume, deng2022gram} to achieve more strict 3D consistency. Among them, methods that directly render their 3D representations for image synthesis achieve the best multiview consistency~\cite{chan2021pi,deng2022gram,skorokhodov2022epigraf,schwarz2022voxgraf,gao2022get3d}. We propose a novel approach for high-quality pose editing of given portraits based on GRAM~\cite{deng2022gram}, which is a recent 3D-aware GAN with state-of-the-art multiview consistency. 
% Despite that, our method can be applied to other 3D-aware-GANs without taking much effort.

\begin{figure*}[t]
	\small
	\centering
	\includegraphics[width=0.99\textwidth]{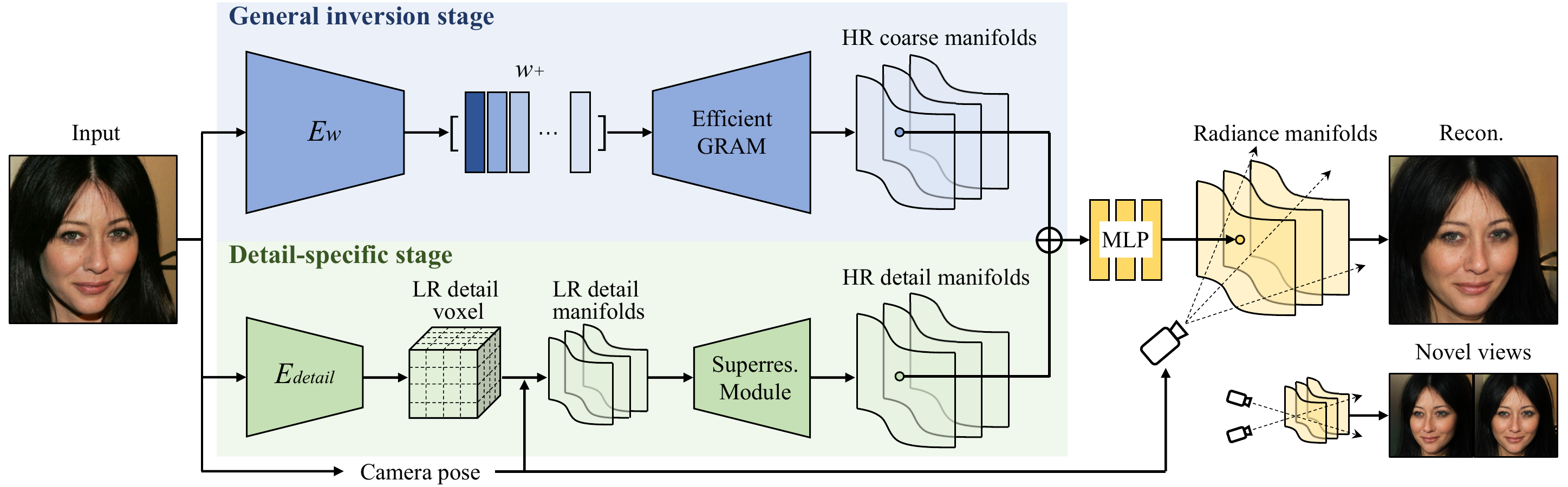}
	\vspace{-5pt}
	\caption{Overview of the GRAMInverter. An input portrait image goes through two stages to obtain the final radiance manifolds for novel view synthesis. The first general inversion stage maps the input image to the latent space of a pre-trained efficient GRAM to obtain coarse radiance manifolds. The second detail-specific stage then extracts detail feature manifolds from the input image and combines them with the coarse results for high-fidelity image synthesis. See the text for more details.}
	\label{fig:framework}
	\vspace{-8pt}
\end{figure*}

\paravspace
\paragraph{GAN inversion.} GAN inversion aims to map a given real image into the latent space of a pre-trained generator for image reconstruction and manipulation. 
% Since the generator is trained to approximate natural image distribution, reasonable manipulations can be obtained via editing the inverted latent code as long as it stays within the meaningful latent distribution of the generator. 
Numerous methods~\cite{zhu2016generative,bau2019seeing,abdal2019image2stylegan, zhu2020domain,karras2020analyzing,richardson2021encoding,tov2021designing,alaluf2021restyle,roich2021pivotal,wang2022high,yao2022feature} try to find a latent code which can faithfully reconstruct the given image meanwhile falls inside a semantically meaningful latent space that supports reasonable editing. They either adopt optimization-based approach~\cite{zhu2016generative,abdal2019image2stylegan,abdal2020image2stylegan++,karras2020analyzing}, introduce an extra image encoder~\cite{richardson2021encoding,tov2021designing,alaluf2021restyle}, or utilize a hybrid version of the former two~\cite{bau2019seeing,zhu2020domain}. Nevertheless, recent studies~\cite{zhu2020improved,tov2021designing,roich2021pivotal,wang2022high} reveal that it is difficult to achieve high-fidelity reconstruction and artifacts-free editing at once given only low-bitrate latent code as representation. As a result, several methods propose to further fine-tune the pre-trained generator~\cite{roich2021pivotal} or allow more detailed features from the input images to leak into the generator during inversion~\cite{wang2022high,yao2022feature}. 

While the above methods target at the inversion of 2D GANs, inverting a given image with 3D-aware GANs shares a similar spirit. An advantage of 3D-aware GAN inversion compare to its 2D counterpart is a natural disentanglement of the 3D pose. Once inverted, novel view synthesis can be easily achieved without further latent space exploration~\cite{shen2020interpreting,harkonen2020ganspace}. The majority of existing 3D-aware GAN inversion methods~\cite{chan2021pi,deng2022gram,chan2021efficient,lin20223d,sun2022fenerf,sun2022ide,wang2022generative,tewari2022disentangled3d} leverage optimization-based or a hybrid approach for faithful reconstruction, which are time-consuming and hard to scale-up. A recent method~\cite{cai2022pix2nerf} explores GAN inversion with a single forward of an encoder, yet it struggles to preserve fine image details. Our proposed method is also an encoder-based inversion approach which yields high-quality reconstruction and novel view synthesis thanks to our novel design.

\paravspace
\paragraph{Pose editing of monocular portraits.} Editing the camera pose of a monocular portrait for novel view synthesis is a longstanding task and has witnessed the emergence of diverse methods. 
% Existing works can be divided into three major categories. 
Some of them~\cite{chai2015high,zhu2015high,zhu2016face,xu2020deep,wu2020unsupervised,pan20202d} achieve pose editing by first conducting 3D reconstruction and then rendering the obtained mesh at novel views. Due to the imperfect reconstruction results, they often have difficulties handling non-face regions and unseen parts at the input view. Others~\cite{wiles2018x2face,siarohin2019animating,siarohin2019first,nirkin2019fsgan,zakharov2020fast,wang2021one,zhou2021pose,ren2021pirenderer,doukas2021headgan,drobyshev2022megaportraits} generate novel views in a face-reenactment paradigm, where warping flows are often learned from video data to transform a source image to a target viewpoint. These methods may encounter geometry distortions at novel views due to the lack of an explicit 3D constraint. More recently, plenty of works~\cite{shen2020interpreting,abdal2021styleflow,tewari2020pie,deng2020disentangled,mallikarjun2021photoapp,hong2022headnerf,cai2022pix2nerf,sun2022fenerf,sun2022ide} have studied pose editing of an image by inverting it into a prior model such as GAN. With the strong prior bearing in a pre-trained generator, synthesizing novel views of a given portrait can be achieved without any 3D or video data during training. Among them, methods using 3D-aware GANs~\cite{deng2022gram,chan2021efficient,cai2022pix2nerf,sun2022fenerf,sun2022ide} shows better 3D consistency under pose variations. Our method is also based on 3D-aware GAN and largely improves the efficiency, reconstruction quality, as well as 3D consistency. 

%--------------------------------------------------------------

\section{Approach}
Given a monocular portrait image $\hat{I}$, we aim to synthesize its novel views at some arbitrary camera viewpoints by leveraging the prior knowledge of a pre-trained 3D-aware GAN, as shown in Fig.~\ref{fig:framework}. 
% \begin{equation}
% 	G:(\bm z, \bm \theta)\in\mathbb R ^{d_z}\times \mathbb R^{3}\rightarrow I\in\mathbb R^{H\times W\times 3}.
% \end{equation}
To guarantee high-quality and 3D-consistent novel view synthesis, we adopt GRAM~\cite{deng2022gram} as our underlying image generator and design an efficient version of it that requires much less computation and memory cost so as to incorporate it into our whole framework (Sec.~\ref{sec:gram}). With the efficient GRAM, we first utilize a \textit{general encoder-based GAN inversion} to reconstruct the coarse radiance manifolds from the input image (Sec.~\ref{sec:general}). We then introduce a \textit{detail-specific reconstruction stage} to learn high-resolution detail manifolds that cannot be well captured by the coarse result, via our proposed novel detail manifolds reconstructor (Sec.~\ref{sec:detail}). Multiple losses are enforced to regulate the predicted detail manifolds to ensure reasonable synthesized results at novel views, by leveraging the 3D priors derived from the coarse radiance manifolds (Sec.~\ref{sec:train}). We describe each part in details below.

\subsection{Efficient Generative Radiance Manifolds}\label{sec:gram}
% We first give a brief review of GRAM originally proposed by~\cite{deng2022gram}, and then introduce our improved version upon it for efficient image generation. 
We start with a brief review of the original GRAM proposed in~\cite{deng2022gram}.
The core of GRAM is its underlying radiance manifolds representation, which regulates radiance field learning on a set of surface manifolds in the 3D space instead of predicting it in the whole volumetric space as done by~\cite{mildenhall2020nerf}. The surface manifolds are defined as a set of iso-surfaces $\{ \mathcal{S}_i \}$ in a 3D scalar field represented by a light-weight MLP called the manifold predictor $\mathcal{M}$:
\begin{equation}
\begin{split}
\vspace{-2pt}
    \mathcal{M}:\bm x \in \mathbb{R}^3 \rightarrow s \in \mathbb{R}, ~\mathcal{S}_i = \{\bm x|\mathcal M(\bm x)=l_i \},
    \end{split} \label{eq:manifold}
\vspace{-2pt}
\end{equation}
where $\{l_i \}$ are $N$ predefined scalar levels. During image generation, only intersections $\{\bm x_i \}$ between a viewing ray $\bm r$ and the surface manifolds will be sent into an MLP-based radiance generator $\Phi$ for radiance prediction:
\begin{equation}
	\Phi:(\bm z,\bm x_i)\in\mathbb{R}^{d_z}\times\mathbb{R}^{3}\rightarrow(\bm c, \alpha)\in\mathbb R^4, \label{eq:radiance}
\end{equation}
where $\bm z$ is a latent code determining the radiance, $\bm c$ is the color, and $\alpha$ is the occupancy. The final color of each ray can be computed via manifold rendering~\cite{deng2022gram,zhou2018stereo}:
\begin{equation}
\begin{split}
\vspace{-3pt}
	C({\bm r}) 
% 	&= \sum_{i=1}^{N}T({\bm x}_i)\alpha({\bm x}_i)\bm c({\bm x}_i) \\
	=   \sum_{i=1}^{N}\prod_{j<i}(1-\alpha({\bm x}_j))\alpha({\bm x}_i)\bm c({\bm x}_i). \label{eq:render}
\vspace{-3pt}
\end{split}
\end{equation}

% The radiance manifold representation helps GRAM to generate high-quality images of virtual subjects with strong 3D consistency. However, its problem is a high memory and computation cost. During training, GRAM requires over $20$GB memory to render a single image of $256\times256$ resolution. This prevents it to be incorporated into our GAN inversion framework with extra image encoders.

The high momery cost of GRAM lies in its MLP-based radiance generator $\Phi$, which requires millions of forward steps to generate a single image. Inspired by the recent EG3D~\cite{chan2021efficient}, we substitute the original radiance generator with a tri-plane generator~\cite{chan2021efficient} based on StyleGAN2 structure~\cite{karras2020analyzing}. Its efficient coarse-to-fine structure helps to reduce memory and computation costs by a large margin. Given the new radiance generator, the color and occupancy for points on the surface manifolds can be obtained by first generating tri-plane features by a 2D CNN $\Psi$, and then conducting tri-plane sampling and sending the sampled features into a small MLP-based decoder $m$ as done in~\cite{chan2021efficient}. Note that although we take the tri-plane generator from EG3D to improve efficiency, we do not use its 2D super-resolution module but keep strictly to the radiance manifolds representation. This helps us to maintain the strong 3D consistency brought by the manifold rendering. In addition, we calculate ray-manifold intersections at $1/4$ resolution of the final image to further speed up our image generation process (see Sec.~\ref{sec:intersection} for details).
% For the manifold predictor $\mathcal{M}$ in GRAM, since it is relatively small, we leave it unchanged. Nevertheless, directly calculating intersections via $\mathcal{M}$~\cite{deng2022gram} for an image at final resolution is still inefficient. Observing that the learned surface manifolds by GRAM on human faces are smooth and nearly planar, we therefore calculate the intersections at $1/4$ resolution of the final image, and obtain the intersections at final resolution via bilinear upsampling. This further improves the efficiency of our image generation process.

The efficient GRAM serves as a strong prior for generating realistic multiview images of virtual subjects. By combining it with our two-stage manifolds reconstruction method, we achieve high-quality novel view synthesis of real portraits, as described in the following sections.

\subsection{General Inversion Stage}\label{sec:general}
% Several previous methods~\cite{chan2021pi,deng2022gram,chan2021efficient,tewari2022disentangled3d} have tried 3D-aware GAN inversion for pose editing of a real portrait, yet they can hardly avoid the tedious optimization step in order for enough reconstruction fidelity. To tackle this problem, we design a two-branch image encoder that allows single-shot high-fidelity pose editing of a portrait image. It consists of a basic branch encoder and a detail branch encoder as shown in Fig.~\ref{fig:framework}.
Given a pre-trained efficient GRAM following a typical 3D-aware GAN training paradigm~\cite{deng2022gram}, we first introduce an image inverter $E_{w}$ that maps a given image to the latent space of the efficient GRAM, as shown in Fig.~\ref{fig:framework}. Inspired by previous StyleGAN-based inversion methods~\cite{abdal2019image2stylegan,tov2021designing,wang2022high}, we invert the given image into a latent code $\bm w+ = [\bm w_1,\bm w_2,...,\bm w_L]$ in $\mathcal{W}+$ space~\cite{abdal2019image2stylegan} of the tri-plane generator $\Psi$ for a proper trade-off between inversion fidelity and pose editing quality, where $L$ is the number of layers in $\Psi$'s synthesis sub-network. We leverage the e4e encoder~\cite{tov2021designing} as the backbone of $E_w$. Given $\bm w+$, we can obtain a coarse radiance manifolds $\Phi(\bm w+, \{\mathcal{S}_i\}) = m\circ\Psi(\bm w+, \{\mathcal{S}_i\})$ via Eq.~\eqref{eq:manifold} and \eqref{eq:radiance}, and further obtain a coarse inversion image $I_w$ by rendering the radiance manifolds at input viewpoint $\hat{\bm\theta}$ via Eq.~\eqref{eq:render}, where $\hat{\bm\theta}$ can be obtained by off-the-shelf 3D face reconstruction method~\cite{deng2019accurate}.

We fixed the pre-trained efficient GRAM and learn the image inverter $E_w$ following the training process of~\cite{tov2021designing}, except that we replaced the adversarial loss in~\cite{tov2021designing} with a naive L2 loss between predicted $\bm w+$ and the average latent code of the $\mathcal{W}+$ space. To further improve the reconstruction fidelity, we fixed the trained $E_w$ and finetuned the efficient GRAM via the pivot tuning strategy~\cite{roich2021pivotal} using all training images. Details for the above training processes can be found in Sec.~\ref{sec:more_training}. 

After training, the general inversion stage can already synthesize reasonable multiview images of the input, yet it cannot faithfully preserve the fine details, making the inverted result looks less like the original image (see Fig.~\ref{fig:ablation}). Therefore, we introduce a detail-specific stage for faithful detail reconstruction, as described below.

\subsection{Detail-Specific Reconstruction Stage} \label{sec:detail}
The detail-specific stage aims to extract fine details from the input image that cannot be well described by the coarse radiance manifolds to improve the reconstruction fidelity. The intuition is to learn high-resolution details in 3D space so that their combination with the coarse radiance manifolds still remains strong 3D consistency under pose variations.
To achieve this goal, we design a detail manifolds reconstructor consisting of two modules: a detail encoder $E_{detail}$ that extracts low-resolution feature voxel from the input image, and a super-resolution module $\mathcal{U}$ to predict high-resolution detail manifolds from the low-resolution voxel. 

\begin{figure}[t]
	\small
	\centering
	\includegraphics[width=1.0\columnwidth]{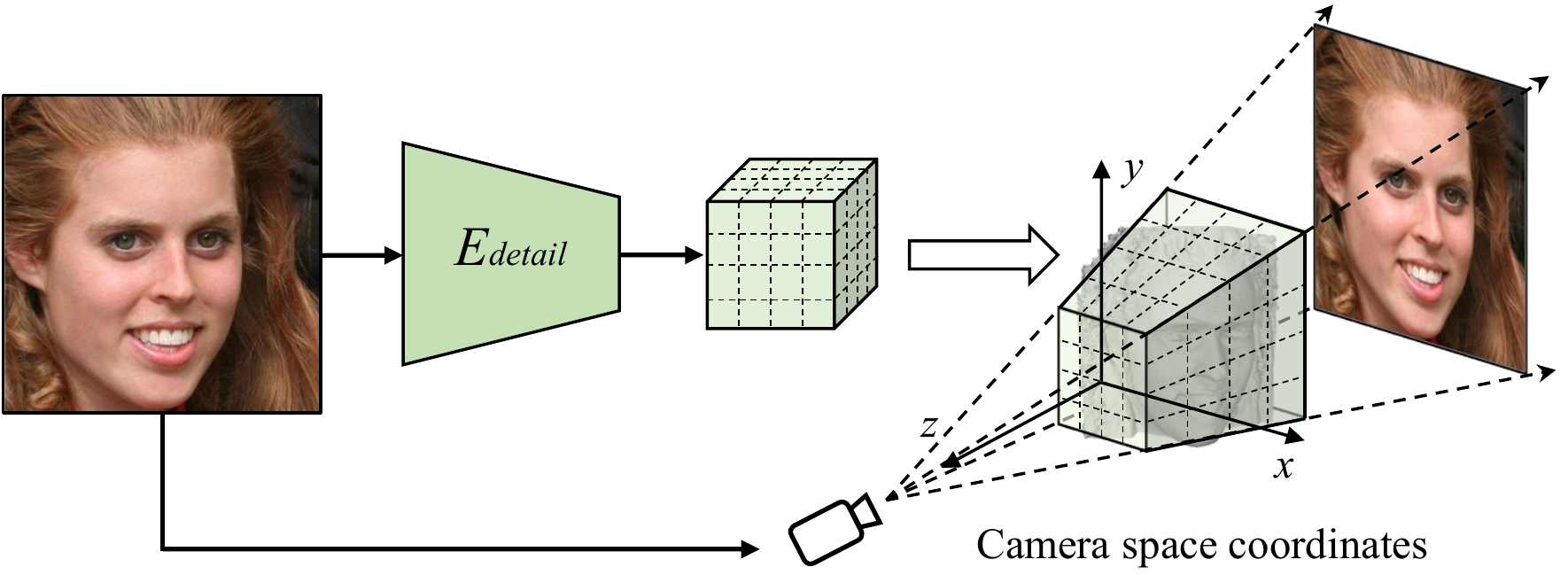}
% 	\vspace{-1pt}
	\caption{The detail encoder $E_{detail}$ extracts a camera space feature voxel from an input image. The voxel corresponds to a quadrangular frustum in the 3D space bounded by the near and far planes of the camera and the outermost viewing rays.}
	\label{fig:detail_voxel}
	\vspace{-7pt}
\end{figure}

Specifically, $E_{detail}$ takes an image $\hat{I}$ as input\footnote{In practice, we find that concatenating $\hat{I}$ with an extra difference map $\Delta = \hat{I} - I_w$ as input yields better reconstruction quality, where $I_w$ is the inversion image obtained by the general inversion stage.} and predicts a camera space feature voxel as shown in Fig.~\ref{fig:detail_voxel}:
\begin{equation}
	E_{detail}: \hat{I} \in \mathbb{R}^{H\times W\times 3} \rightarrow V \in \mathbb{R}^{H_{lr}\times W_{lr} \times D_{lr} \times d_V},
\end{equation}
where $d_V$ is the feature dimension. We implement $E_{detail}$ as a 3D U-Net with skip connections to extract both global geometry structures as well as local fine textures from the input image. 
We refer the readers to Sec.~\ref{sec:structure} for detailed network structure. 

The feature voxel is defined in camera space instead of world space (\ie space where tri-plane features of efficient GRAM are defined) as it is easier for $E_{detail}$ to extract image-aligned features than to learn transformed world space features (see Sec.~\ref{sec:ablation}). 
% In Sec.~\ref{sec:experiment}, we show that this greatly improves fine details preservation after inversion. 
Given the feature voxel $V$, we can obtain the corresponding feature $f^{lr}\in \mathbb{R}^{d_V}$ for a point $\bm x\in\mathbb{R}^3$ in the world space via:
\begin{equation}
\vspace{-1pt}
f^{lr} = {\rm grid\_sample}(V,{\rm world2cam}(\bm x)), \label{eq:query}
\vspace{-1pt}
\end{equation}
where ${\rm grid\_sample}$ is a tri-linear interpolation function, and ${\rm world2cam}$ is a transformation function between the world space and the camera space.

Nevertheless, since $V$ is a low-resolution voxel, directly combining it with the feature manifolds obtained from the general inversion stage leads to a blurry inversion result, while predicting a high-resolution voxel (\eg $256^3$) instead causes unaffordable memory cost. Inspired by~\cite{xiang2022gram}, we take the advantage of our radiance manifolds representation to obtain high-resolution detail manifolds from the low-resolution voxel via manifold super-resolution. Specifically, we first obtain low-resolution detail manifolds $f^{lr}(\{\mathcal{S}_i\})$ by querying features from $V$ via Eq.~\eqref{eq:query} for low-resolution points grid on the surface manifolds $\{\mathcal{S}_i\}$. We then flatten each manifold to a low-resolution feature map $F^{lr}_i$ and send it to the super-resolution module $\mathcal{U}$ to obtain a high-resolution feature map $F^{hr}_i = \mathcal{U}(F^{lr}_i)$, where $\mathcal{U}$ is a simple 2D CNN of $4$ convolution blocks and $2$ bilinear upsampling blocks. Finally, we obtain the high-resolution detail manifolds $f^{hr}(\{\mathcal{S}_i\})$ by re-projecting each flattened feature map $F^{hr}_i$ to the surface manifolds. Since we conduct super-resolution for 3D space surface manifolds, 3D consistency across different views can be naturally maintained during this process. Note that although the manifold super-resolution strategy is proposed in~\cite{xiang2022gram}, it does not leverage it in a reconstruction scenario but to generate random fine details. By contrast, we utilize it for faithful reconstruction of high-frequency details in the original image. 

Given $f^{hr}(\{\mathcal{S}_i\})$ from the detail-specific stage, we add it to the coarse feature manifolds $\Psi(\bm w+, \{\mathcal{S}_i\})$ from the general inversion stage, and send each feature point on the manifolds to the MLP-based decoder $m$ to obtain the final radiance manifolds, as shown in Fig.~\ref{fig:framework}. The final inversion image $I$ can then be obtained similarly via manifold rendering at input view $ \hat{\bm \theta}$. Novel views can also be easily generated given an arbitrary camera pose $\bm \theta$ during rendering.

\subsection{Detail Manifolds Learning}\label{sec:train}
% We learn the efficient GRAM ($\mathcal{M}$, $\Psi$ and $m$) and the two-branch image encoder ($E_w$, $E_{detail}$ and $\mathcal{U}$) on in-the-wild monocular images by a multi-stage training strategy.

% We first learn the efficient GRAM following a typical 3D-aware GAN training paradigm as in~\cite{deng2022gram}. 
% Then, we fixed the pre-trained efficient GRAM and learn the basic branch encoder $E_w$ following the training process of~\cite{tov2021designing}, except that we replaced the adversarial loss in~\cite{tov2021designing} with a naive L2 loss between predicted $\bm w+$ and the average latent code of the $\mathcal{W}+$ space. To further improve the reconstruction fidelity of the basic branch, we fix $E_w$ from the last stage and finetune the efficient GRAM via the pivot tuning strategy~\cite{roich2021pivotal} using all training images. Details for the above training stages can be found in the \emph{suppl. material}.

We fix the image inverter $E_w$ and the efficient GRAM from the general inversion stage, and learn the detail manifolds reconstructor with the following losses.

\begin{figure}[t]
	\small
	\centering
	\includegraphics[width=0.95\columnwidth]{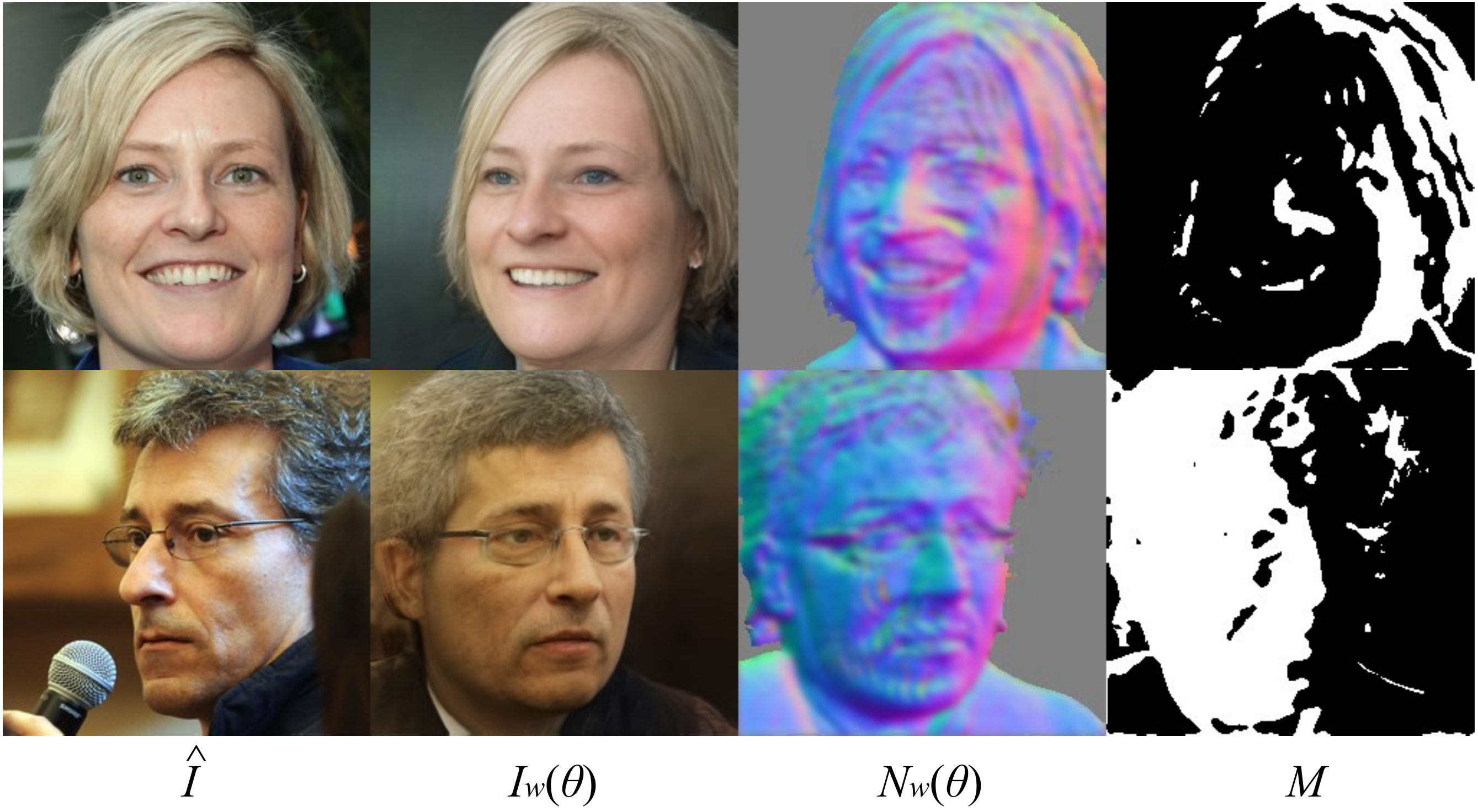}
	\caption{Visualization of the novel view regularization.}
	\label{fig:view_loss}
	\vspace{-7pt}
\end{figure}

\paravspace
\paragraph{Image reconstruction loss.} A multi-level reconstruction loss is applied between the final inversion image $I$ and the input image $\hat{I}$:
\vspace{-2pt}
\begin{equation}
\resizebox{0.9\hsize}{!}{$\mathcal{L}_{r} = ||I-\hat{I}||^2 + {\rm LPIPS}(I,\hat{I}) + (1 - \langle f_{id}(I),f_{id}(\hat{I})\rangle)$,}
\label{eq:recon}
\end{equation}
where ${\rm  LPIPS}(\cdot,\cdot)$ is the perceptual loss defined by~\cite{zhang2018unreasonable}, and $f_{id}$ is a pre-trained face recognition network~\cite{deng2019arcface}.

\paravspace
\paragraph{Novel view regularization.} The reconstruction loss guarantees a faithful inversion result at the input viewpoint, yet artifacts can still occur when rendering the radiance manifolds at other views. We, therefore, design a regularization term to ensure reasonable novel view synthesis results:
\vspace{-2pt}
\begin{equation}
\mathcal{L}_{nv} = {\rm LPIPS}(M\odot I(\bm \theta),M\odot I_w(\bm \theta)),
\label{eq:view}
\vspace{-2pt}
\end{equation}
where $I(\bm \theta)$ and $I_w(\bm \theta)$ are final and coarse inversion image rendered at novel view $\bm \theta$ respectively, $\odot$ is element-wise multiplication, and $M$ is a binary mask:
\vspace{-2pt}
\begin{equation}
M(u,v) = \mathbb{I}(-\bm r(\hat{\bm \theta})\cdot{N_w(\bm \theta)(u,v)}<\tau).
\label{eq:mask}
\vspace{-2pt}
\end{equation}
Here $(u,v)$ is the image space coordinate, $\mathbb{I}$ is the indicator function, $\bm r(\hat{\bm \theta})$ is the camera lookat direction of the input image $\hat{I}$, $N_w(\bm \theta)$ is the surface normal map of $I_w(\bm \theta)$, and $\tau$ is a scalar threshold. The intuition behind this regularization is that the details for regions unobserved in the input image should stay close to the coarse inversion result at novel views, as the coarse inversion image is more reasonable at new views due to leveraging the priors from the pre-trained GRAM. The normal map $N_w(\bm \theta)$ in Eq.~\eqref{eq:mask} can be effectively calculated via the following equation:
\vspace{-2pt}
\begin{equation}
N_w(\bm \theta)(u,v) = -\frac{1}{\eta}\sum_{i=1}^{N}T(\alpha({\bm x}_i))\alpha({\bm x}_i)\frac{\partial \alpha({\bm x}_i)}{\partial {\bm x}_i },
\label{eq:normal}
\vspace{-3pt}
\end{equation}
where ${\bm x}_i$ are intersections along the ray that $(u,v)$ corresponds to, $T(\alpha({\bm x}_i)) = \prod_{j<i}(1-\alpha({\bm x}_j))$ is the accumulated transparency, and $\eta$ is a normalizing scalar. The partial gradient $\partial \alpha({\bm x}_i)/\partial {\bm x}_i $ can be easily computed via backpropagating the MLP-based decoder $m$. Visualizations of the normal map $N_w(\bm \theta)$ and the binary mask $M$ are in Fig.~\ref{fig:view_loss}.

\paravspace
\paragraph{Depth regularization.} We further enforce a depth regularization to the HR detail manifolds $f^{hr}(\{\mathcal{S}_i\})$ to ensure that details are predicted near the geometry surface:
\begin{equation}
\mathcal{L}_{depth} = \left\{
\begin{aligned}
& \lambda \|f^{hr}(\bm x_i)\|^2 & |z(\bm x_i)-z_{surf}|>\epsilon\\
& 0 & |z(\bm x_i)-z_{surf}|\leq\epsilon
\end{aligned}
\right.,
\label{eq:depth}
\end{equation}
where $\bm x_i$ are intersections along the viewing rays at input viewpoint $\hat{\bm \theta}$, $z(\bm x_i)$ is the depth of $\bm x_i$, $z_{surf}=\sum_{i=1}^{N}T(\alpha({\bm x}_i))\alpha({\bm x}_i)z(\bm x_i)$ is the depth of the approximated surface, and $\epsilon$ is a threshold. This regularization ensures correct parallax for the learned details (see Sec.~\ref{sec:ablation}).

% \begin{figure}[t]
% 	\small
% 	\centering
% 	\includegraphics[width=1.0\columnwidth]{zoomin.pdf}
% 	\caption{Dolly zoom effect of a portrait image.}
% 	\label{fig:zoom}
% \end{figure}

\section{Experiments}\label{sec:experiment}
\paragraph{Implementation details.} We train our method on the FFHQ~\cite{karras2019style} dataset at $256\times256$ resolution and test it on the CelebA-HQ~\cite{karras2017progressive} dataset. All images are pre-processed following the procedure in~\cite{deng2022gram}. The camera pose of input images is estimated by the face reconstruction method of~\cite{deng2019accurate}. We train our models on $4$ NVIDIA Tesla V100 GPUs with $32$GB memory. The whole training process takes around $6$ days, where training the efficient GRAM takes $2$ days, training the image inverter and finetuning the efficient GRAM takes $2$ days and $1$ day respectively, and training the detail manifolds reconstructor takes $1$ day. See Sec.~\ref{sec:implement} for more details.

% \begin{table}[t]
%     \centering
%     \caption{}       \label{tab:inversion_compare}
%     \begin{tabular}{c|c|c|c|c|c}
%     \toprule[1pt]
%     Methods & \!PSNR $\uparrow$\! & \!SSIM $\uparrow$\! & \!LPIPS $\downarrow$ \!& \!ID $\uparrow$\! & \!FID $\downarrow$\!\\
%     \hline
%     e4e & 19.23 & 0.451 & 0.213 & 0.639 & 35.92\\
%     HFGI & \textbf{22.30} & 0.579 & 0.135 & 0.779 & \textbf{26.41}\\
%     \hline
%     \!pix2NeRF\! &15.75 & 0.358 & 0.482 & 0.299 & \\
%     IDE-3D& 16.73 & 0.382 & 0.290 &0.231&51.51\\
%     Ours & 21.46 & \textbf{0.649} &\textbf{0.128} &\textbf{0.927} & 28.17\\
%     \bottomrule[1pt]
%     \end{tabular}
% \end{table}

\subsection{Novel View Synthesis Results}
Figure~\ref{fig:teaser} shows the novel view synthesis results of our method given different portrait images. Our method well preserves fine details (\eg hair bangs, wrinkles, moles) of the input images and produces their 3D consistent novel views. 
% Figure~\ref{fig:zoom} further shows an example of dolly zoom effect. Thanks to our underlying 3D representation, this can be easily achieved by simultaneously varying the camera fov and the viewing distance. 
The whole inversion and novel view synthesis process runs at $3$ FPS on a V100 GPU without specialized acceleration, which largely improves the efficiency upon previous optimization-based 3D-aware GAN inversions. With the manifold caching technique in~\cite{deng2022gram}, we can further increase the free view rendering speed to 180FPS. More results are in Sec.~\ref{sec:more_results}.

\subsection{Comparison with Prior Arts} \label{sec:compare}
\paragraph{Comparison with GRAM.} We first compare our efficient version of GRAM with the original one~\cite{deng2022gram}. We measure the image generation quality by the Fr\'echet Inception Distances (FID)~\cite{heusel2017gans} between $20$K randomly generated images and $20$K sampled real images. The 3D consistency is measured by the reconstruction quality of NeuS~\cite{wang2021neus} (\ie PSNR$_{mv}$ and SSIM$_{mv}$) on multiview images of $50$ generated instances following~\cite{xiang2022gram}. As shown in Tab.~\ref{tab:gram}, our efficient GRAM largely reduces the memory cost and increases the inference speed upon the original one without sacrificing image generation quality or 3D consistency, by introducing the StyleGAN2-based radiance generator and the efficient intersection calculation strategy. This improvement enables our following GRAMinverter method, otherwise, it is difficult, if not impossible, to leverage the memory-consuming GRAM for encoder-based GAN inversion. We also list the performance of the state-of-the-art EG3D~\cite{chan2021efficient} as a reference. Although EG3D has better image quality, it sacrifices the 3D consistency which we argue is a key factor for 3D-aware generation.

\begin{table}[t]
% \multicolumn{5}{c}{\multirow{3}{*}{}}
    \centering
    \small
    \caption{Comparison between efficient GRAM and GRAM~\cite{deng2022gram}. *: Inference on a Tesla V100 GPU with a batchsize of $1$.} \label{tab:gram}
    \vspace{-2pt}
    \begin{tabular}{l|ccccc}
    \toprule[1pt]
   \! Methods\! & \!\!Memory* $\downarrow$\!\! & \!\!FPS* $\uparrow$\!\! & \!\!FID $\downarrow$\!\!& \!\!\!PSNR$_{mv}$ $\uparrow$\!\!\! & \!\!\!SSIM$_{mv}$ $\uparrow$\!\!\!\\
    \hline
    \!EG3D\!& 2.8G & 20 & 6.02& 34.0 & 0.928\\
    \hline
    \!GRAM\!& 12G & 2 & 15.0& \textbf{38.0} & 0.966\\
%   \!\!GRAM-HD\!\!& & & 13.0 & 36.5 & 0.955\\
    \!Ours\!& \textbf{3.3G} & \textbf{14} & \textbf{14.2} & 37.6 & \textbf{0.969} \\

    \bottomrule[1pt]
    \end{tabular} 
\vspace{-8pt}
\end{table}

\begin{table*}[t]
% \multicolumn{5}{c}{\multirow{3}{*}{}}
    \centering
    \small
    \caption{Quantitative comparison with existing portrait editing methods. See the text for details.}       \label{tab:edit_compare}
    \vspace{-2pt}
    \begin{tabular}{l|ccccc|cc|cc}
    \toprule[1pt]
    &\multicolumn{5}{c|}{Inversion fidelity} & \multicolumn{2}{c|}{\!Novel view quality\!} & \multicolumn{2}{c}{3D consistency} \\
    Methods & \!\!PSNR $\uparrow$\!\! & \!\!SSIM $\uparrow$\!\! & \!\!LPIPS $\downarrow$ \!\!& \!\!ID $\uparrow$\!\! &\!\! FID $\downarrow$ \!\!& \!ID$_{nv}$ $\uparrow$\! & \!FID$_{nv}$ $\downarrow$\! & \!PSNR$_{mv}$ $\uparrow$\! & \!SSIM$_{mv}$ $\uparrow$\!\\
    \hline
    PIRenderer~\cite{ren2021pirenderer} &--&--&--&--&--&0.476&42.64&36.72&0.958\\
    Face-vid2vid~\cite{wang2021one}&--&--&--&--& --& 0.416 &\cellcolor{yellow}41.76&36.10&0.942\\
    \hline
    \!\!e4e~\cite{tov2021designing} + InterfaceGAN~\cite{shen2020interpreting}\!\! & \!\cellcolor{yellow}19.23\! & \!\cellcolor{yellow}0.451\! & \!\cellcolor{yellow}0.213\! &\!\cellcolor{yellow}0.706\!&\!\cellcolor{yellow}35.92\!&\cellcolor{yellow}0.489&\cellcolor{orange}38.04&34.29&0.909\\
    \!HFGI~\cite{wang2022high} + InterfaceGAN~\cite{shen2020interpreting}\!& \!\cellcolor{pink}{22.30}\! & \!\cellcolor{orange}0.579\! & \!\cellcolor{orange}0.135\! & \!\cellcolor{orange}0.827\!& \!\cellcolor{pink}{26.41}\!&\cellcolor{orange}0.516&45.23&33.99&0.917\\
    \!HFGI~\cite{wang2022high} + StyleHEAT~\cite{yin2022styleheat}\!& \!\cellcolor{pink}{22.30}\! & \!\cellcolor{orange}0.579\! & \!\cellcolor{orange}0.135\! & \!\cellcolor{orange}0.827\!& \!\cellcolor{pink}{26.41}\!&\!0.457\!&\!58.33\!&35.93&0.951\\
    \hline
    \!pix2NeRF~\cite{cai2022pix2nerf}\! &\!16.95\! &\! 0.394\! &\! 0.452\! & \!0.466\! & \!108.3\!&0.378&115.6&\cellcolor{pink}50.66&\cellcolor{pink}0.997\\
    IDE-3D (encoder)~\cite{sun2022ide}& \!16.73\! & \!0.382\! &\! 0.290\! &\!0.393\!& \!51.51\!&0.324&47.56&\cellcolor{yellow}37.57&\cellcolor{yellow}0.950\\
    Ours & \!\cellcolor{orange}21.51\! & \!\cellcolor{pink}{0.650}\! &\!\cellcolor{pink}{0.127}\! &\!\cellcolor{pink}{0.936}\! &\!\cellcolor{orange}28.17\!&\!\cellcolor{pink}0.635\!&\!\cellcolor{pink}36.02\!&\cellcolor{orange}39.53&\cellcolor{orange}0.974\\
    \bottomrule[1pt]
    \end{tabular}
% \vspace{-5pt}
\end{table*}

\begin{figure*}[t]
	\small
	\centering
	\includegraphics[width=0.97\textwidth]{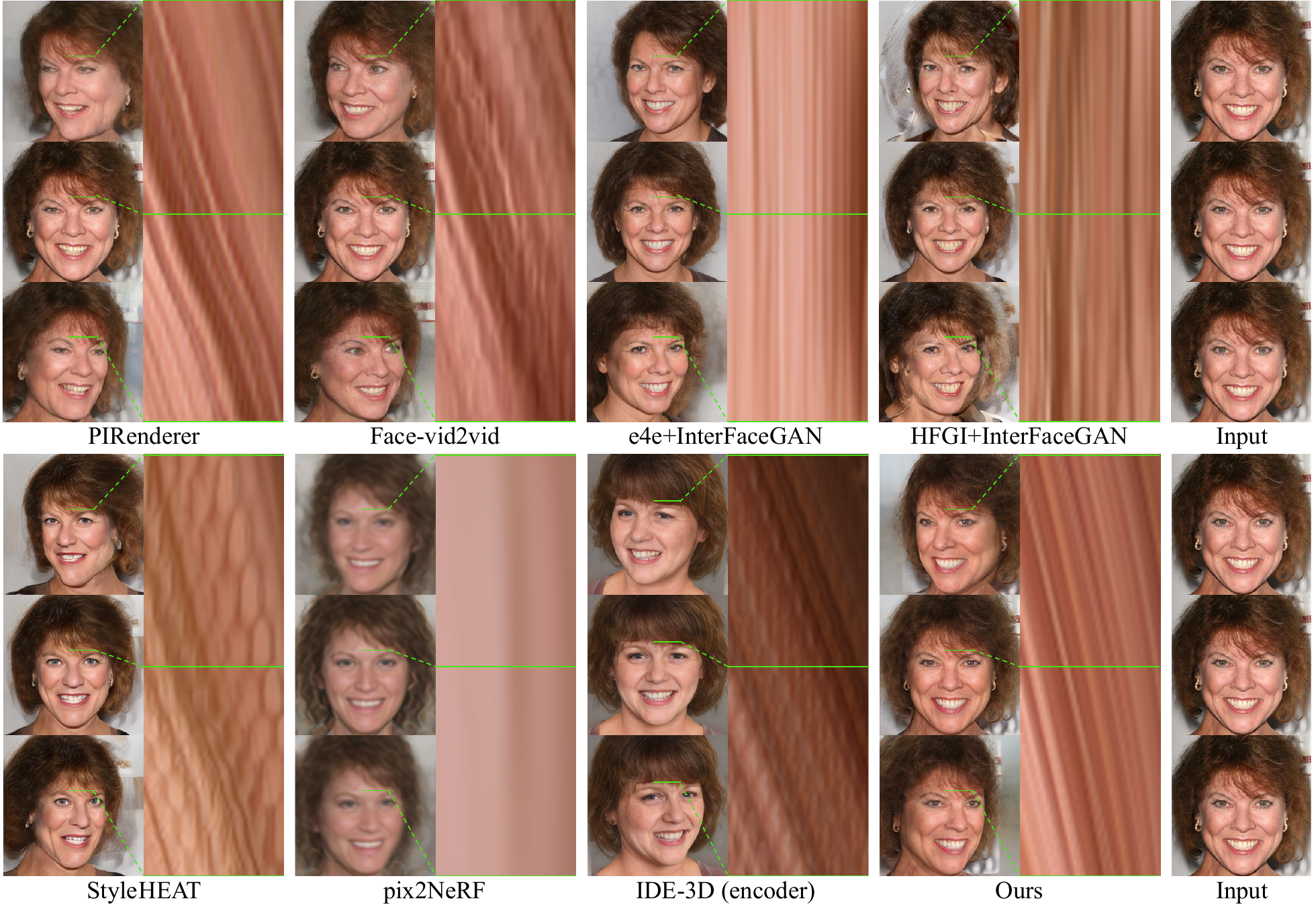}
    \vspace{-5pt}
    	\caption{Pose editing comparison. Texture images with smoothly tilted strips indicate better 3D consistency. \textbf{Best viewed with zoom-in.}}
	\label{fig:compare}
	\vspace{-7pt}
\end{figure*}

\begin{figure}[t]
	\small
	\centering
	\includegraphics[width=1.0\columnwidth]{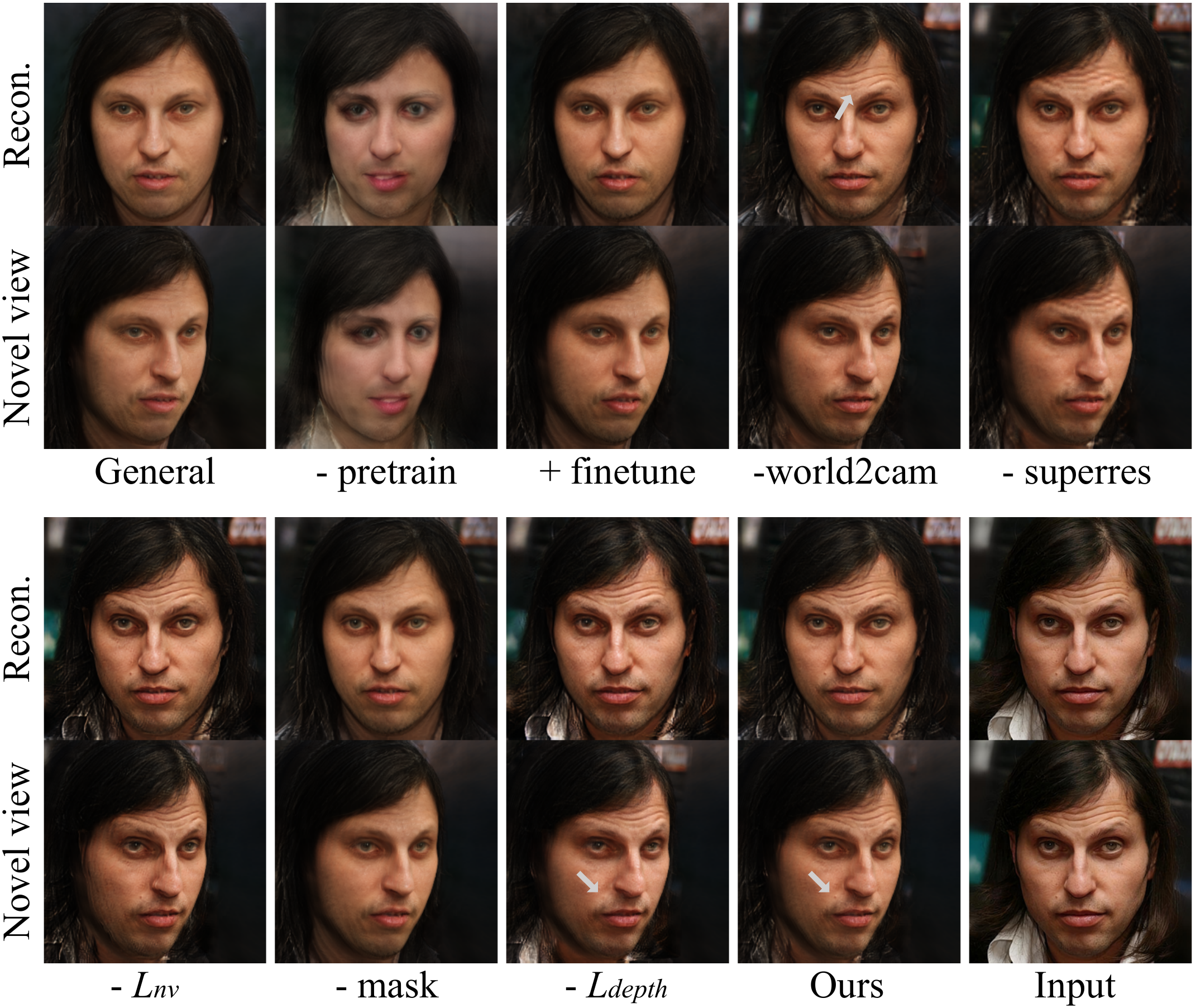}
    \vspace{-10pt}
    	\caption{Visual comparison between different alternatives. Our final solution yields the best result. \textbf{Best viewed with zoom-in.}}
	\label{fig:ablation}
	\vspace{-7pt}
\end{figure}

\begin{table}[t]
% \multicolumn{5}{c}{\multirow{3}{*}{}}
    \centering
    \small
    \caption{Ablation study of our proposed framework.}       \label{tab:ablation}
    \vspace{-1pt}
    \begin{tabular}{l|cccc}
    \toprule[1pt]
    Methods &\! \!PSNR $\uparrow$\!\! & \!\!LPIPS $\downarrow$\! \!&\! \!ID $\uparrow$\!\!& \!\!FID$_{nv}$ $\downarrow$\!\! \\
    \hline
    General & 17.68 & 0.265 & 0.648 & 44.43\\
  General - pretrain & 17.45& 0.280 & 0.472 & 52.12\\
    General + finetune & 18.00 & 0.254 & 0.678 & 43.46\\
    \hline
    Detail (Ours) & \cellcolor{yellow}21.51 & \cellcolor{yellow}0.127 & \cellcolor{yellow}0.936 & \cellcolor{orange}36.02\\
    Detail - world2cam & 19.96 & 0.171 & 0.908 & 38.74\\
    Detail - superres & 21.06 & 0.165 & 0.926 & 38.45\\
    Detail - $\mathcal{L}_{nv}$ & \cellcolor{orange}22.32 & \cellcolor{orange}0.106 & \cellcolor{orange}0.949 & \cellcolor{yellow}36.49\\
    Detail - mask& 19.08&0.211&0.840& 43.25\\
    Detail - $\mathcal{L}_{depth}$ & \cellcolor{pink}23.68 & \cellcolor{pink}0.094 & \cellcolor{pink}0.960 & \cellcolor{pink}35.60\\

    \bottomrule[1pt]
    \end{tabular}
\vspace{-3pt}
\end{table}

\paravspace
\paragraph{Comparison with pose editing methods.} We compare with existing methods that can achieve 3D pose editing of a given portrait via single forward pass, including 2D GAN inversion-based methods: e4e~\cite{tov2021designing}+InterFaceGAN~\cite{shen2020interpreting}, HFGI~\cite{wang2022high}+InterFaceGAN, and HFGI+StyleHEAT~\cite{yin2022styleheat}; 3D-aware GAN inversion-based methods: pix2NeRF~\cite{cai2022pix2nerf} and IDE-3D~\cite{sun2022ide}; and face reenactment methods: PIRenderer~\cite{ren2021pirenderer} and Face-vid2vid~\cite{wang2021one}.

We first evaluate the inversion fidelity among GAN inversion-based methods (Tab.~\ref{tab:edit_compare}). We report PSNR, SSIM, LPIPS, identity similarity (\ie ID) measured by cosine distance of face recognition features~\cite{wang2018cosface}, and FID. All metrics are calculated between the first $1$K images of CelebA-HQ and their corresponding inverted results. Since different methods may generate images of different resolutions and alignments, we pre-process all results following~\cite{deng2022gram} and resize them to $256\times256$ for a fair comparison. As shown, our method significantly outperforms other 3D-aware GAN inversion methods across all metrics. We also exceed the StyleGAN2-based inversion method e4e and achieve comparable results with the state-of-the-art method HFGI.

We further compare our method with other approaches on pose editing of portrait images. We generate novel views (see Sec.~\ref{sec:nv_detail} for details) of the $1$K test images using different methods and evaluate their identity similarity and FID to the original input in Tab.~\ref{tab:edit_compare}. Higher ID$_{nv}$ and lower FID$_{nv}$ indicate that a method can better keep the identity and image quality while changing the camera pose. Our method yields the best result among all competitors. We also surpass PIRenderer and Face-vid2vid which require video data for training, while ours is merely trained on monocular in-the-wild images. Figure~\ref{fig:compare} shows a visual comparison.

Finally, we measure the 3D consistency of all methods during continuous variation of the camera viewpoint. Following~\cite{xiang2022gram}, for each method, we generate $30$ images under different views for $50$ test instances in CelebA-HQ, and measure the multiview reconstruction quality of NeuS on them (\ie PSNR$_{mv}$ and SSIM$_{mv}$). In theory, better 3D consistency across different views would reduce the learning difficulty of NeuS, thus leading to higher PSNR and SSIM. Table~\ref{tab:edit_compare} shows that our method has the second best 3D consistency among all methods, while the best one (\ie pix2NeRF) generates over-smooth images of low quality as shown in Fig.~\ref{fig:compare} and indicated by the high FID score in Tab.~\ref{tab:edit_compare}. Our method outperforms IDE-3D in that it utilizes a 2D super-resolution module in its 3D-aware GAN which lowers the 3D consistency to some extent. Nevertheless, all 3D-aware GAN-based methods yield better 3D consistency compared to other 2D methods, indicating the importance of 3D-aware GAN for pose editing of images. A further comparison with the full pipeline of IDE-3D which includes an extra optimization step is in the Sec.~\ref{sec:more_compare}. 

Figure~\ref{fig:compare} further shows the visual comparison of 3D consistency, where we draw the stacked texture image of a fixed horizontal line segment during continuous camera movement following~\cite{xiang2022gram}. Methods with strong 3D consistency will result in texture images with smoothly tilted strips, while methods with low 3D consistency produce twisted textures (\ie geometry distortions and texture flickering issues) or vertical lines (\ie texture sticking issues). Our method clearly produces a more reasonable texture image compared to the others. See Sec.~\ref{sec:more_compare} for more results.

\subsection{Ablation Study} \label{sec:ablation}
We conduct an ablation study to validate the efficacy of our proposed framework and report the results in Tab.~\ref{tab:ablation} and Fig.~\ref{fig:ablation}. All metrics are calculated similarly as in Sec.~\ref{sec:compare}.

\paravspace
\paragraph{Inversion stage.} Table~\ref{tab:ablation} shows the performance of different stages. \textit{General} stands for the general inversion stage without finetuned generator. \textit{General - pretrain} denotes learning the efficient GRAM with $E_w$ together instead of pre-training it via the 3D-aware GAN framework. \textit{General + finetune} denotes finetuning the pre-trained efficient GRAM as described in Sec.~\ref{sec:general}, and \textit{Detail} denotes our final approach with detail-specific reconstruction. As shown, the general stage alone cannot produce faithful reconstruction result, whether the efficient GRAM is further finetuned or not. By contrast, introducing the detail-specific reconstruction significantly improves the inversion fidelity upon the previous stage without sacrificing novel view quality. Learning the efficient GRAM together with the encoder without pre-training leads to a significant performance drop, indicating the importance of leveraging the prior knowledge of a pre-trained 3D-aware GAN.

\paravspace
\paragraph{Network architecture.}  We ablate the architecture of the detail manifolds reconstructor. As shown in Tab.~\ref{tab:ablation} and Fig.~\ref{fig:ablation}, learning the low-resolution detail voxel in world space instead of in camera space (\textit{- world2cam}) harms the reconstruction fidelity. And removing the super-resolution module for high-resolution manifold prediction (\textit{- superres}) leads to blurry inversion results.

\paravspace
\paragraph{Regularization.} We further validate our proposed regularization for detail manifolds learning. As shown in Tab.~\ref{tab:ablation} and Fig.~\ref{fig:ablation}, removing the novel view regularization (\textit{- $\mathcal{L}_{nv}$}) causes obvious artifacts at new views and leads to the increase of FID$_{nv}$, though it improves the reconstruction quality at the input viewpoint. Simply enforcing $\mathcal{L}_{nv}$ without the normal-aware mask (\textit{- mask})
damages fine texture preservation at visible regions. Finally, although learning without the depth regularization (\textit{- $\mathcal{L}_{depth}$}) results in better metrics, we find that it cannot well preserve certain fine details at novel views due to incorrect parallax brought by the depth error (\eg mole in Fig.~\ref{fig:ablation}). We conjecture that such dynamic artifact can hardly be captured by the current feature extractor~\cite{szegedy2016rethinking} for FID computation.

\section{Conclusions}
We presented GRAMinverter, a novel approach for high-fidelity and 3D-consistent portrait synthesis from monocular images via single forward pass. The core idea is to learn a detail manifolds reconstructor to predict 3D-consistent fine details on the radiance manifolds from a input image, and combine them with the coarse radiance manifolds obtained via an encoder-based inversion of the pre-trained GRAM. Extensive experiments have demonstrated our superior results over previous works. We believe our method paves a new way for efficient 3D-aware portrait creation.

\paravspace\paragraph{Limitations and future works.} Our GRAMinverter has several limitations. Based on the radiance manifold representation, it produces layered artifacts at large viewing angles. It cannot well handle occlusions of hands and other accessories. Its performance is also affected by the training data and may produce inferior results for out-of-distribution input. Besides, it does not support editing of attributes beyond camera viewpoints as done in previous 2D GAN inversions. Better 3D representations and inversion strategies should be further explored to alleviate these problems.
 
%%%%%%%%% REFERENCES
{\small
\bibliographystyle{ieee_fullname}
\bibliography{egbib}
}

\clearpage

\appendix

\begin{strip}
\centering
\Large{\textbf{Supplementary Material}}
\end{strip}

\renewcommand{\thesection}{\Alph{section}}
\renewcommand{\thefigure}{\Roman{figure}}
\renewcommand{\thetable}{\Roman{table}}
\renewcommand{\theequation}{\Roman{equation}}
\setcounter{figure}{0}
\setcounter{equation}{0}

\section{More Implementation Details}\label{sec:implement}

 \subsection{Data Preparation}
 
 We align all images in FFHQ~\cite{karras2019style} and CelebA-HQ~\cite{karras2017progressive} using the detected facial landmarks following~\cite{deng2022gram}. Specifically, we first use an off-the-shelf landmark detector~\cite{bulat2017far} to extract 5 facial landmarks for each image. Then, we resize and crop the images by solving a least square problem between the detected landmarks and canonical 3D landmarks from the average shape of a 3D face model~\cite{paysan20093d}. Camera poses of the images are extracted using a 3D face reconstruction model~\cite{deng2019accurate}.
 
 \subsection{Network Structure}\label{sec:structure}
\begin{figure*}[p]
    \small
	\centering
	\includegraphics[width=0.95\textwidth]{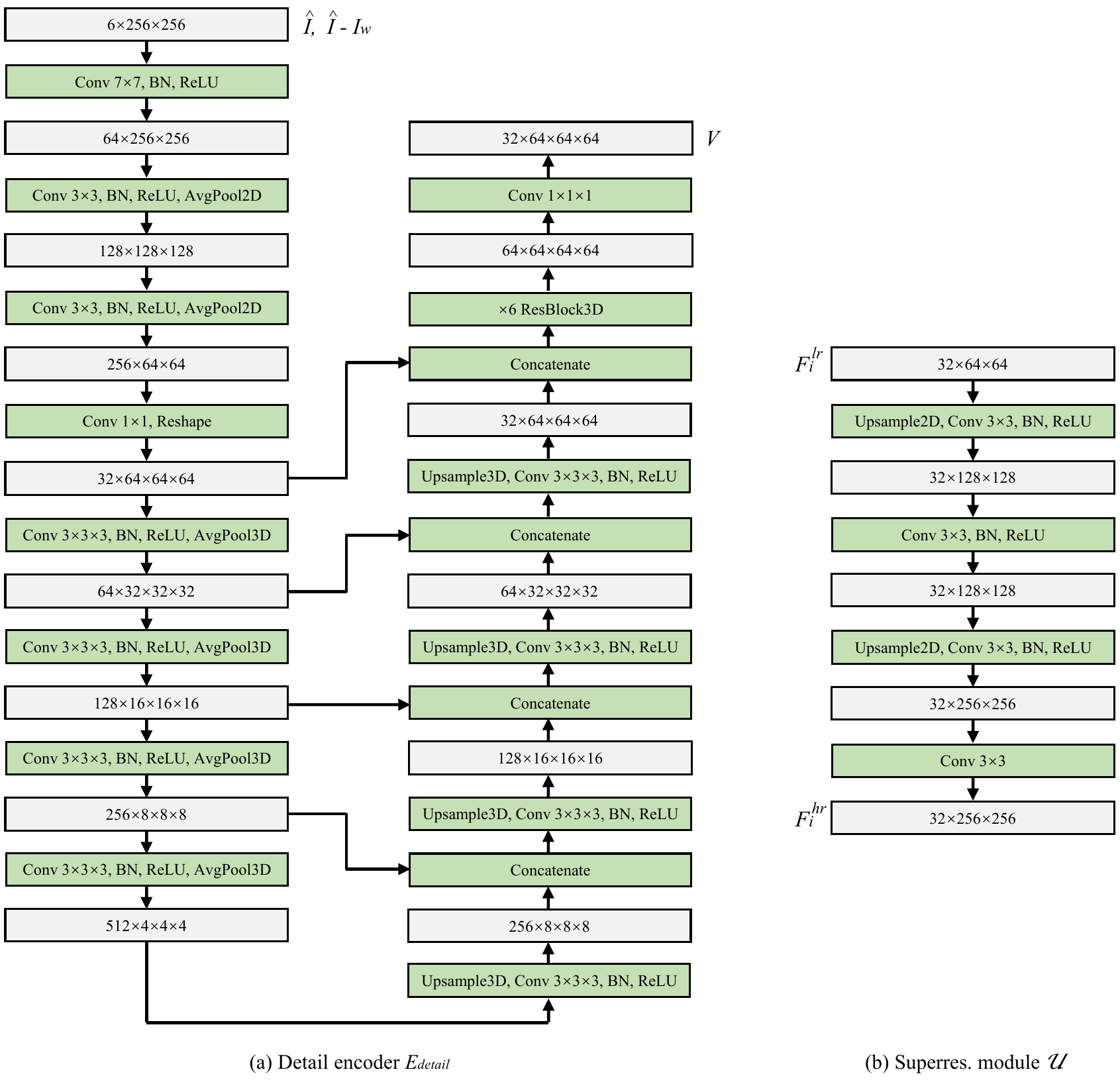}
	
	\caption{Network structure of the detail manifolds reconstructor. It consists of a detail encoder $E_{detail}$ and a super-resolution module $\mathcal{U}$.\label{fig:network}}
	
% 	\vspace{-8pt}
\end{figure*}

 The structure of the detail manifolds reconstructor is shown in Fig.~\ref{fig:network}. It consists of two sub-networks. A detail encoder $E_{detail}$ and a super-resolution module $\mathcal{U}$.

 \paragraph{Detail encoder $E_{detail}$.} The detail encoder receives the concatenation of the input image $\hat{I}$ and the difference map $\hat{I} - I_w$, and predicts a low-resolution feature voxel $V$ (see Fig.~\ref{fig:network}~(a)). It consists of several 2D downsampling blocks, followed by a 2D convolution to project the low-resolution 2D feature map to 3D voxel. A 3D U-Net structure with skip connections is then applied, followed by several 3D resblocks to obtain the final low-resolution feature voxel $V$.

  \paragraph{Super-resolution module $\mathcal{U}$.} The super-resolution module takes the low-resolution feature map $F_i^{lr}$ derived from each low-resolution feature manifold as input, and produces a high-resolution feature map $F_i^{hr}$ which will be later projected back to the surface manifolds (see Fig.~\ref{fig:network}~(b)). It consists of two upsampling blocks, and each block contains two 2D convolutions.

 \subsection{Intersection Calculation Details}\label{sec:intersection}
\begin{figure}[t]
	\small
	\centering
	\includegraphics[width=1.0\columnwidth]{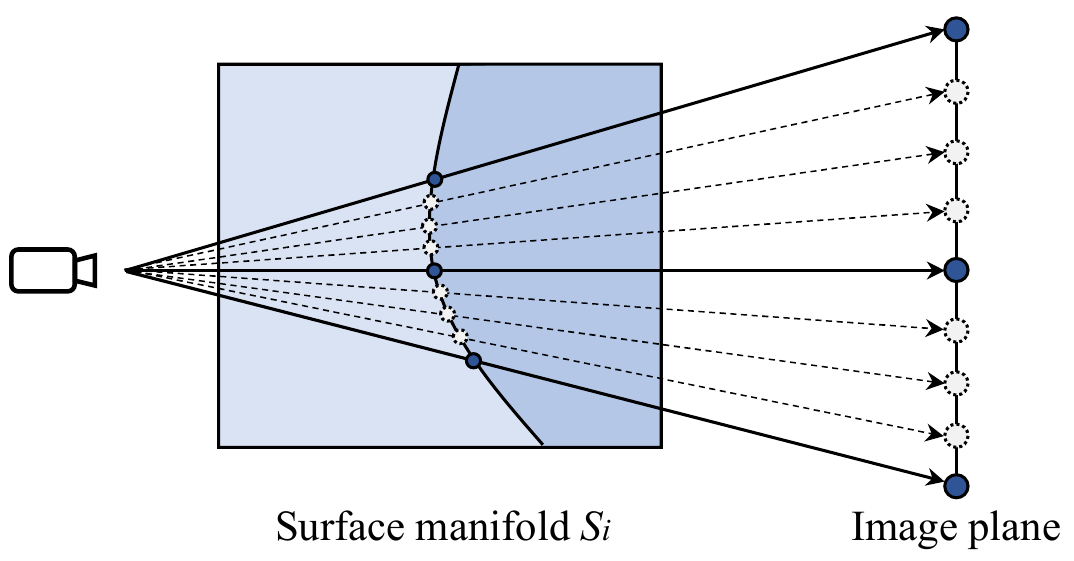}
	\vspace{-10pt}
	\caption{Illustration of our intersection calculation. Ray-manifold intersections are first calculated at $1/4$ resolution of the final image (\ie blue dots), and then go through bilinear upsampling to obtain dense intersections at the final resolution (\ie gray dots).}
	\label{fig:intersection}
	\vspace{-8pt}
\end{figure}

The efficient GRAM requires to calculate ray-manifold intersections for manifold rendering following~\cite{deng2022gram}. To accelerate the efficiency of this process, we calculate ray-manifold intersections at $1/4$ resolution of the final image, as depicted in Fig.~\ref{fig:intersection}. 

Specifically, we first generate viewing rays at a resolution of $64\times64$, and calculate their intersections with each surface manifold produced by the manifold predictor $\mathcal{M}$ following~\cite{deng2022gram}. Then, we upsample the obtained low-resolution intersection grid on each manifold via bilinear interpolation to obtain dense intersections at the final resolution (\ie $256\times$256). In this way, only the low-resolution intersections obtained in the first step require forwarding the manifold predictor, which largely reduces the computation cost compare to directly calculating intersections at the final resolution. Since the learned surface manifolds for human faces have small curvature and are nearly planar at local regions (see illustration in~\cite{deng2022gram}), the intersections obtained via the bilinear upsampling are close to the ground truth and have a minor influence on the final synthesis results.

 \subsection{More Training Details} \label{sec:more_training}
 
 \paragraph{Pretraining efficient GRAM.} We follow~\cite{deng2022gram} to train the efficient GRAM on FFHQ dataset at a resolution of $256\times256$. During training, we randomly sample latent code $\bm z$ from the normal distribution and camera pose $\bm \theta$ from the estimated distribution of the training data and send them to the efficient GRAM to generate corresponding images. The manifold predictor $\mathcal{M}$ is initialized following~\cite{deng2022gram}. The tri-plane generator $\Psi$ and the MLP-based decoder $m$ are initialized following~\cite{chan2021efficient}.
 The synthesized images, together with randomly sampled real images from the training data, are sent into an extra discriminator~\cite{karras2020analyzing} for loss computation. We adopt the non-saturating GAN loss with R1 regularization~\cite{mescheder2018training} to learn the efficient GRAM and the discriminator. We also enforce the pose regularization in~\cite{deng2022gram} to ensure that the learned geometries are reasonable.
 
 We use the Adam optimizer~\cite{kingma2015adam} with $\beta_1 = 0$ and $\beta_2=0.99$. The learning rates are set to $2.5e-3$ for the tri-plane generator and the MLP-based decoder, $2e-5$ for the manifold predictor, and $1e-3$ for the discriminator. The loss weights for the R1 regularization and the pose regularization are set to $10$ and $30$, respectively. We trained the efficient GRAM for $150$K iterations with a batchsize of $32$. Training took 2 days on $4$ NVIDIA Tesla V100 GPUs with $32$GB memory.
 
%  \paravspace
 \paragraph{General inversion stage.} During this stage, we fix the pre-trained efficient GRAM and learn the image inverter $E_w$. The image inverter is initialized following~\cite{tov2021designing}. 
 We adopt the multi-level reconstruction loss $\mathcal{L}_r$ in Eq.~\eqref{eq:recon} for faithful image reconstruction, and the minimal variation loss $\mathcal{L}_{d-reg}$ proposed in~\cite{tov2021designing} to ensure that each $\bm w_i, i= 1,...,L$ in the predicted $\bm w^+$ latent code are close to each other.
 Besides, we apply a regularization on the predicted latent code $\bm w^+$ to ensure that it falls in a semantically meaningful latent space:
 \begin{equation}
 \mathcal{L}_{\bm w+} = ||\bm w^+ - \bar{\bm w}^+ ||^2,
 \end{equation}
 where $\bar{\bm w}^+$ is the average latent code of the $\mathcal{W}+$ space computed using $10$K randomly sampled $\bm z$.
 
We use the Adam optimizer with $\beta_1 = 0.9$ and $\beta_2=0.999$. The initial learning rate for the image inverter is $3e-4$, and decreases to $6e-5$ after $100$K iterations. The balancing weights for the three terms in $\mathcal{L}_r$ are set to $1e-2$, $1$, and $4e-2$, respectively. The weights for $\mathcal{L}_{d-reg}$ and $\mathcal{L}_{\bm w+}$ are $1e-3$ and $1e-4$, respectively. The network is trained for $150$K iterations with a batchsize of $32$, which took $2$ days on $4$ NVIDIA Tesla V100 GPUs.

 \paragraph{General inversion stage - finetuning.} After the image inverter is learned, we further finetune the efficient GRAM for better image reconstruction. We only finetune the tri-plane generator $\Psi$ and the MLP-based decoder $m$, and leave the manifold predictor $\mathcal{M}$ unchanged. The two networks are learned following~\cite{roich2021pivotal}. Specifically, we adopt the multi-level reconstruction loss $\mathcal{L}_r$ in Eq.~\eqref{eq:recon}. In addition, we leverage the locality regularization $\mathcal{L}_R$ proposed in~\cite{roich2021pivotal} to ensure that images synthesized by the finetuned efficient GRAM stay close to those of the original one at randomly sampled locations in the latent space. Different from \cite{roich2021pivotal}, we finetune the efficient GRAM on the whole training set instead of using only a single image.
 
 During training, the Adam optimizer is also applied with $\beta_1 = 0.9$ and $\beta_2=0.999$. The learning rate for the efficient GRAM is $1e-3$. The balancing weights for $\mathcal{L}_r$ are similar to the above stage, and the weight for $\mathcal{L}_R$ is set to $0.5$. We use a batchsize of $16$ and finetune the efficient GRAM for $100$K iterations. The whole process took $1$ day on $4$ NVIDIA Tesla V100 GPUs.
 
 \paragraph{Detail-specific reconstruction stage.} Finally, we fix the image inverter as well as the efficient GRAM learned from the previous stages, and learn the detail manifolds reconstructor via the losses proposed in Sec.~\ref{sec:train}. We set the balancing weights for $\mathcal{L}_r$ following the above stages. The loss weights for the novel view regularization $\mathcal{L}_{nv}$ and the depth regularization $\mathcal{L}_{depth}$ are set to $4$ and $2e-4$, respectively. We use the Adam optimizer with $\beta_1 = 0.9$ and $\beta_2=0.999$, and set the learning rate for the detail manifolds reconstructor to $3e-4$. We use a batchsize of $8$ and train the whole pipeline for $60$K iterations. It took $1$ day on $4$ NVIDIA Tesla V100 GPUs.
 
 \subsection{Baseline Implementation Details}
 \paragraph{PIRenderer.} PIRenderer~\cite{ren2021pirenderer} is a face-reenactment method learned on video data. It leverages a 3D Morphable Model (3DMM)~\cite{paysan20093d,deng2019accurate} as guidance and learns 2D warping flow to drive a source image with target motions. It supports intuitive control of a given image by directly modifying the input 3DMM parameters to the network.
 We use the officially released code and model trained on VoxCeleb~\cite{nagrani2017voxceleb} dataset\footnote{\href{https://github.com/RenYurui/PIRender}{https://github.com/RenYurui/PIRender}} in our experiments, and achieve pose editing of an image by modifying the input 3D pose parameters.
 
  \paragraph{Face-vid2vid.} Face-vid2vid~\cite{wang2021one} is also a face-reenactment method learned on video data. It extracts 3D keypoints of an image and derives 3D warping flows from them to transfer the 3D features of a source image to a target position. By using a single frame as both the source and the target, and applying 3D rotation to the extracted 3D keypoints of the target, it can also achieve intuitive control over the 3D pose of a given portrait image. Since the official code and model are unavailable, we use a re-implementation of it trained on VoxCeleb dataset\footnote{\href{https://github.com/zhanglonghao1992/One-Shot\_Free-View\_Neural\_Talking\_Head\_Synthesis}{https://github.com/zhanglonghao1992/One-Shot\_Free-View\_Neural\_Talking\_Head\_Synthesis}} for our experiments.
 
 \paragraph{e4e.} e4e~\cite{tov2021designing} is an encoder-based StyleGAN2~\cite{karras2020analyzing} inversion method. Its encoder adopts a feature-pyramid structure~\cite{lin2017feature} and predicts StyleGAN2's $\mathcal{W}+$ space vector for a given image. By editing the predicted latent code towards certain direction, and sending the modified code into the pre-trained StyleGAN2, it can achieve pose control of the given image. We use the official released code and model trained on FFHQ\footnote{\href{https://github.com/omertov/encoder4editing}{https://github.com/omertov/encoder4editing}} to carry out our experiments.
  
 \paragraph{HFGI.} HFGI~\cite{wang2022high} is also an encoder-based StyleGAN2 inversion method. It  builds upon the e4e method and extracts extra feature maps from a given image as substitutions to the original feature maps within StyleGAN2. Therefore, it achieves more faithful inversion results compare to e4e. We use its officially released code and model trained on FFHQ\footnote{\href{https://github.com/Tengfei-Wang/HFGI}{https://github.com/Tengfei-Wang/HFGI}} in our experiments.

 \paragraph{InterFaceGAN.} InterFaceGAN~\cite{shen2020interpreting} is a latent space editing method for StyleGAN~\cite{karras2019style} and StyleGAN2. It learns the binary classification boundaries of multiple image attributes for latent vectors in StyleGAN's $\mathcal{W}+$ space. By modifying the latent code along the direction perpendicular to an interface, it can change the corresponding attribute of a synthesized image. It can also be combined with GAN inversion methods like e4e and HFGI for real image editing. Since the officially released model only contains interfaces for StyleGAN, we use the model provided by~\cite{wang2022high} for StyleGAN2-based pose editing.

 \paragraph{StyleHEAT.} StyleHEAT~\cite{yin2022styleheat} is also a latent space editing method for StyleGAN2 which targets at talking head synthesis. Different from InterFaceGAN, it modifies the latent feature maps within the StyleGAN2 instead of the $\mathcal{W}+$ space latent vector. It learns 2D warping flows for the feature maps via the help of video data as well as the guidance of 3DMM, similarly as done by PIRenderer. It also supports direct 3D pose editing of a given image by modifying the 3D pose parameters for generating the warping flow. We use the officially released code and model trained on VoxCeleb\footnote{\href{https://github.com/FeiiYin/StyleHEAT/}{https://github.com/FeiiYin/StyleHEAT/}} in our experiments.

 \paragraph{pix2NeRF.} pix2NeRF~\cite{cai2022pix2nerf} is an encoder-based 3D-aware GAN inversion method based on pi-GAN~\cite{chan2021pi}. It simultaneously learns an image encoder and a 3D-aware image generator to reconstruct NeRF~\cite{mildenhall2020nerf} representation from a given image for novel view synthesis. We adopt its official released code\footnote{\href{https://github.com/primecai/Pix2NeRF}{https://github.com/primecai/Pix2NeRF}} in our experiments. Since the official model is trained on CelebA~\cite{liu2015deep} dataset that overlaps with our test set, we re-train it on the FFHQ dataset for a fair comparison.

 \paragraph{IDE-3D.} IDE-3D~\cite{sun2022ide} is a 3D-aware GAN aiming for 3D-consistent portrait synthesis with interactive control. Its image generator is based on the tri-plane generator and the 2D super-resolution module proposed in~\cite{chan2021efficient}. It also achieves disentangled editing of real images by introducing a hybrid GAN inversion scheme, where it first learns an image encoder to map a given image into the latent space of the pre-trained generator, and then leverages instance-specific optimization~\cite{roich2021pivotal} to further improve the reconstruction fidelity. We use its official model trained on FFHQ\footnote{\href{https://github.com/MrTornado24/IDE-3D}{https://github.com/MrTornado24/IDE-3D}} in our experiments. Moreover, we only use the inversion results from its encoder instead of those from the further optimization step for a fair comparison. A comparison with its full pipeline including the optimization step is demonstrated in Tab.~\ref{tab:compare_ide3d} and Fig.~\ref{fig:compare_ide3d}.

   \subsection{Visualization Details}
  Visualization results in this paper are rendered with yaw angles ranging from $-0.4$ rad to $0.4$ rad. The pitch angles are identical to those of the input images, which are estimated via the face reconstruction method of~\cite{deng2019accurate}. The roll angles are set to zero.
 
 \subsection{Novel View Experiment Details}\label{sec:nv_detail}
 We describe more details about the novel view synthesis comparison proposed in Sec.~\ref{sec:compare}. Specifically, we generate novel views of the first $1$K test images in the CelebA-HQ dataset using different methods to calculate the metrics (\ie ID$_{nv}$ and FID$_{nv}$ in Tab.~2). We randomly sample the yaw angle within a range of $[-0.5,-0.4]\cup[0.4,0.5]$, and set the pitch and roll identical to those of the original input. To ensure that the novel view images have a large pose difference with the input, we multiply the sampled yaw angle by $-1$ if its absolute difference with that of the input is smaller than $0.3$. For all methods, we use the same $1$K sampled yaw angles to generate novel view images for a fair comparison. 

  \begin{table}[t]
% \multicolumn{5}{c}{\multirow{3}{*}{}}
    \centering
    \small
    \caption{Comparison with the full pipeline of IDE-3D which contains an extra optimization step.}\label{tab:compare_ide3d}
    \vspace{-10pt}
    \begin{tabular}{l|ccccc}
    \toprule[1pt]
   \! Methods\! & \!\!\!PSNR $\uparrow$\!\!\!& \!\!\!LPIPS $\downarrow$\!\!\!& \!\!\!ID$_{nv}$ $\uparrow$\!\!\! & \!\!\!PSNR$_{mv}$ $\uparrow$\!\!\!& \!\!\! Time(s) $\downarrow$  \!\!\!\\
    \hline
    \!\!\!IDE-3D (full)\!\!\!& \textbf{24.43} & \textbf{0.092} & 0.507 & 37.10 & 100 \\
%   \!\!GRAM-HD\!\!& & & 13.0 & 36.5 & 0.955\\
    \!\!Ours\!\!& 21.57 &0.123 & \textbf{0.645} & \textbf{39.53} & \textbf{0.3}\\

    \bottomrule[1pt]
    \end{tabular} 
\vspace{-2pt}
\end{table}
\begin{figure}[t]
	\small
	\centering
	\vspace{-2pt}
	\includegraphics[width=1.0\columnwidth]{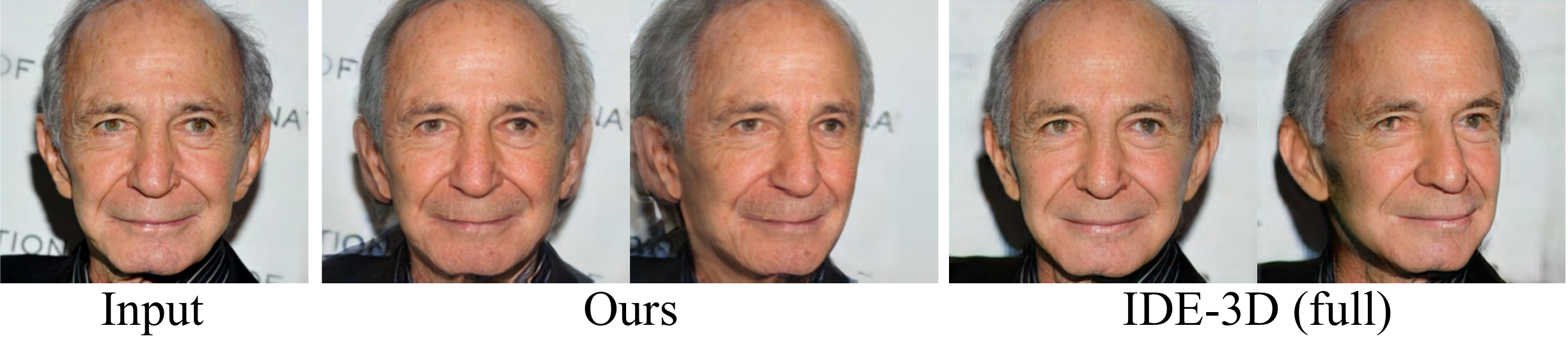}
	\vspace{-15pt}
	\caption{Comparison with the full pipeline of IDE-3D.}\label{fig:compare_ide3d}
\end{figure}
 
\section{More Results} \label{sec:more_results}
\subsection{Novel View Synthesis Results}
\begin{figure*}[p]
	\small
	\centering
	\includegraphics[width=1.0\textwidth]{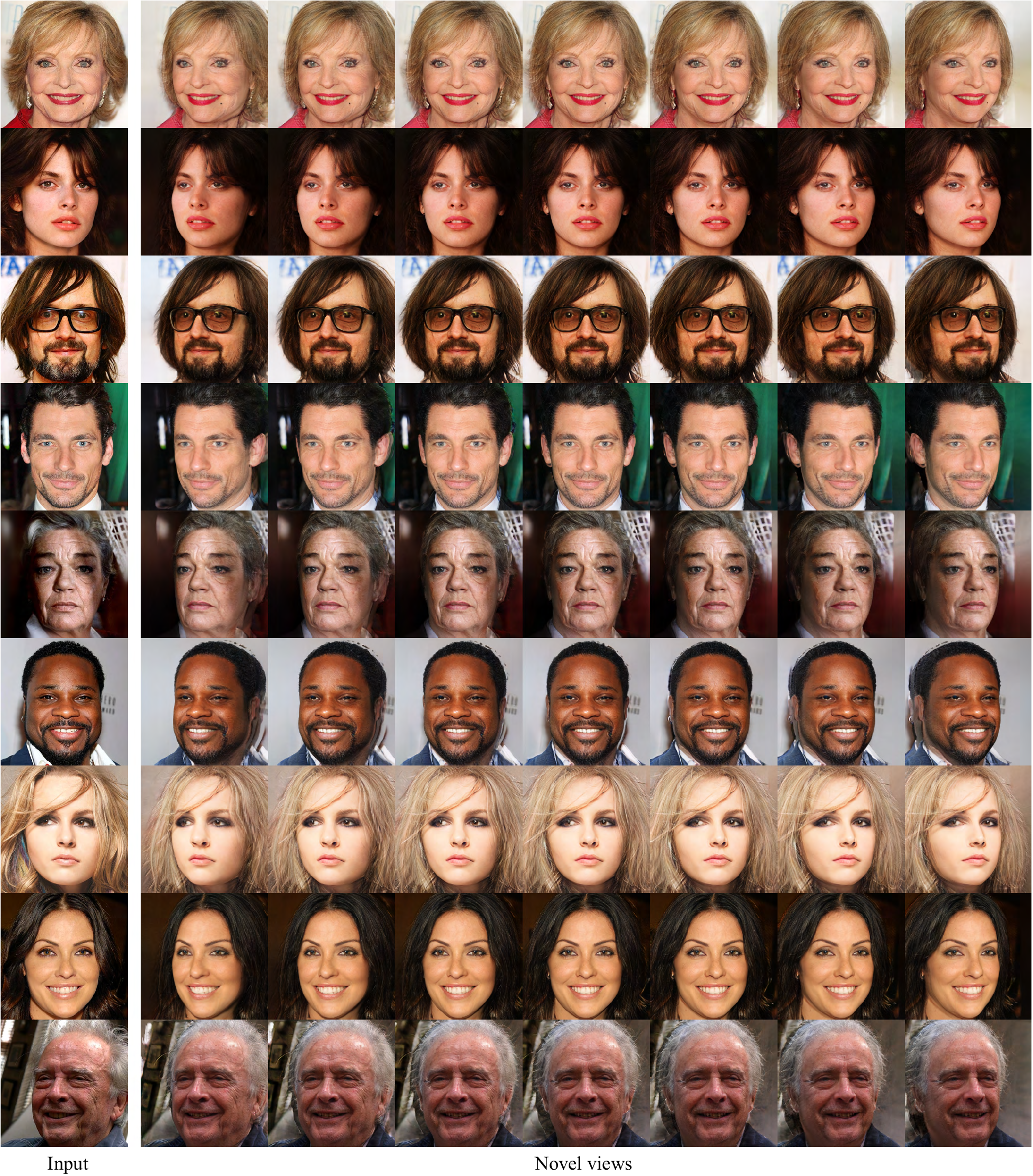}
	\caption{More novel view synthesis results on CelebA-HQ by our method. \textbf{Best viewed with zoom-in.}}
	\label{fig:more_result1}
\end{figure*}

\begin{figure*}[p]
	\small
	\centering
	\includegraphics[width=1.0\textwidth]{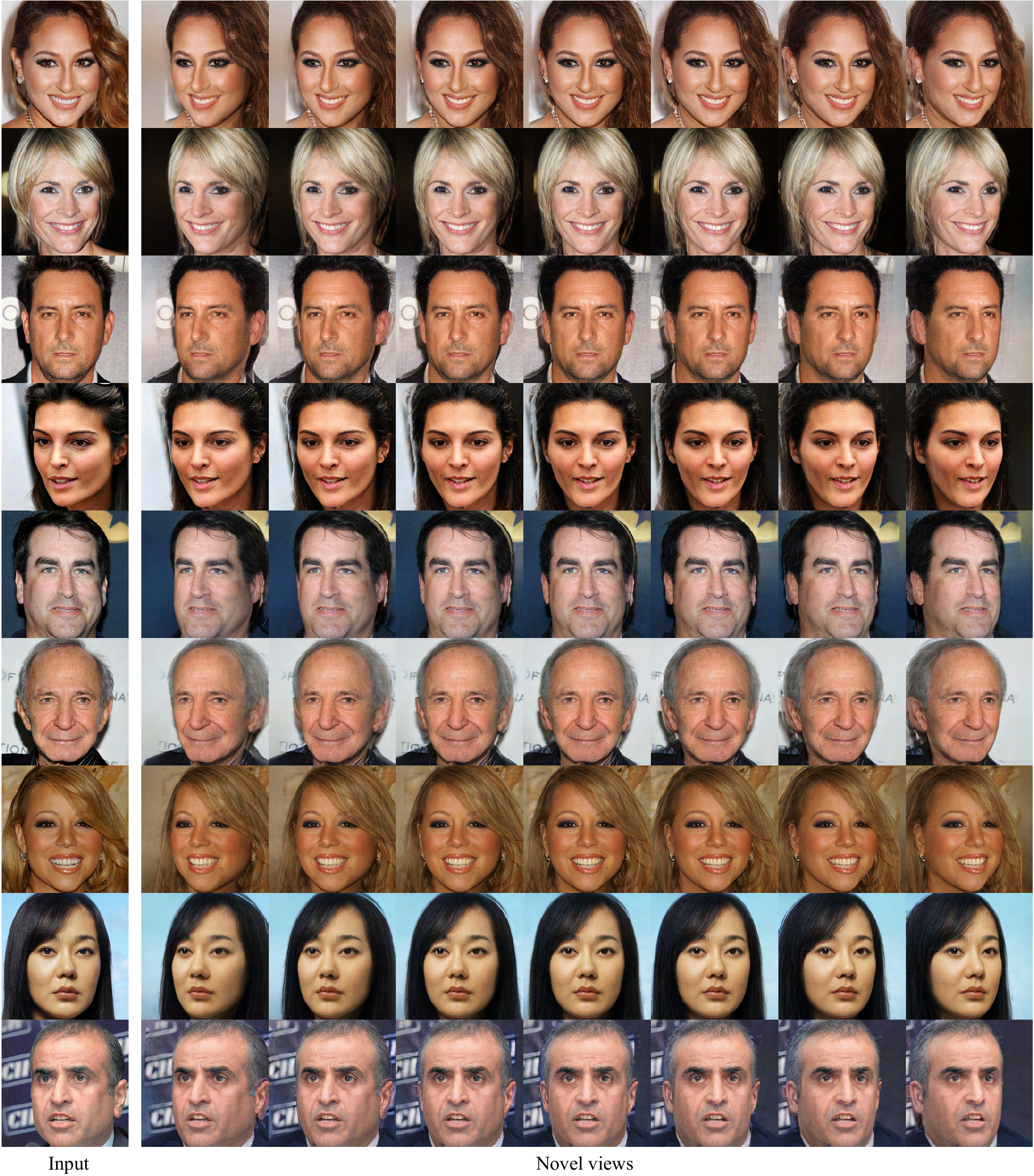}
	\caption{More novel view synthesis results on CelebA-HQ by our method. \textbf{Best viewed with zoom-in.}}
	\label{fig:more_result2}
\end{figure*}

\begin{figure*}[p]
	\small
	\centering
	\includegraphics[width=1.0\textwidth]{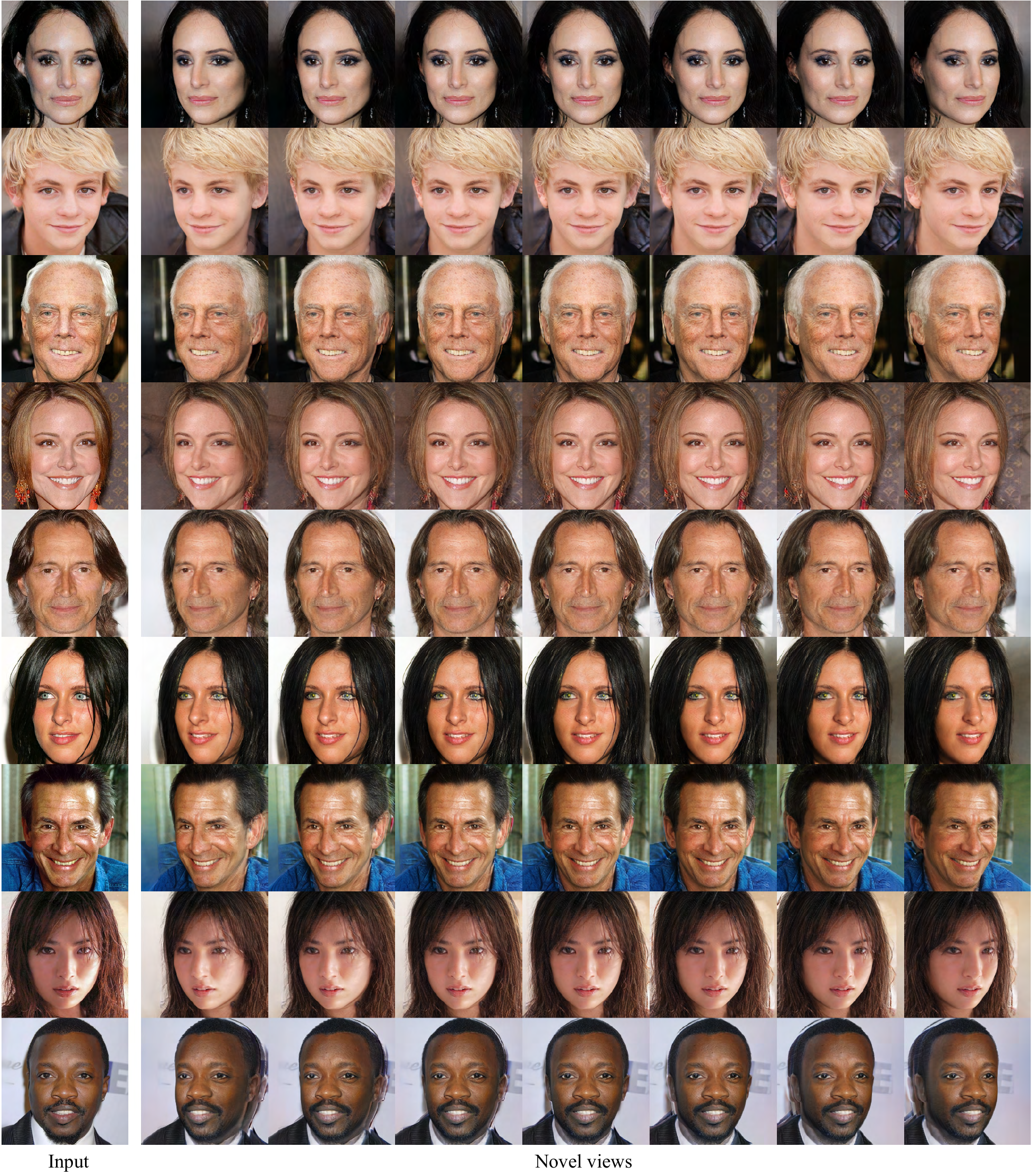}
	\caption{More novel view synthesis results on CelebA-HQ by our method. \textbf{Best viewed with zoom-in.}}
	\label{fig:more_result3}
\end{figure*}

\begin{figure*}[p]
	\small
	\centering
	\includegraphics[width=1.0\textwidth]{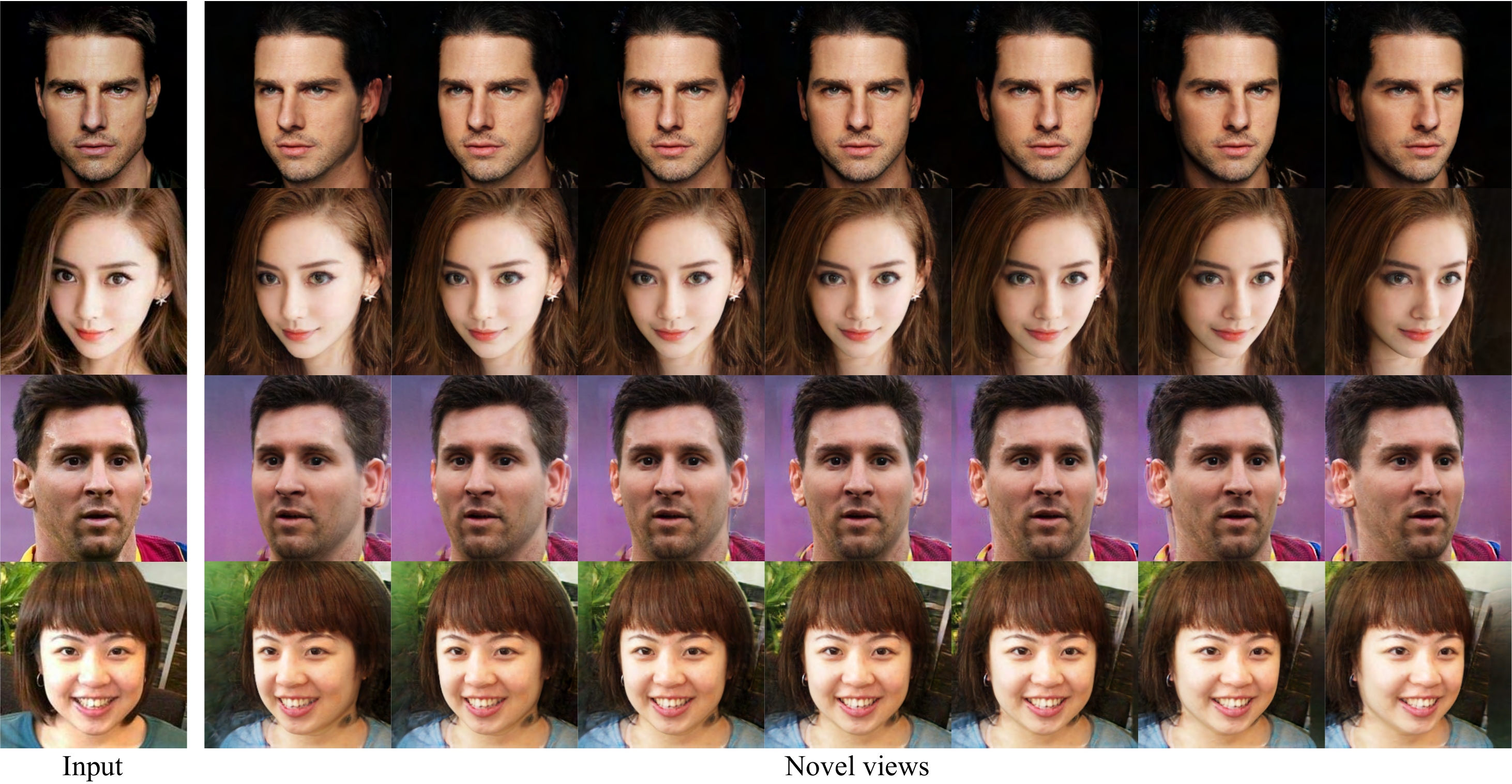}
	\caption{More novel view synthesis results on in-the-wild images. \textbf{Best viewed with zoom-in.}}
	\label{fig:wild}
\end{figure*}

Figure~\ref{fig:more_result1}, \ref{fig:more_result2} and \ref{fig:more_result3} shows more novel view synthesis results by our method on CelebA-HQ test data. Figure~\ref{fig:wild} further shows novel view synthesis results on in-the-wild images. Our method can generate realistic novel views with high fidelity and strong 3D consistency for diverse subjects. \textbf{Please see \href{https://yudeng.github.io/GRAMInverter}{project page} for animations.}

\subsection{Comparisons with the Prior Art} \label{sec:more_compare}
\begin{figure*}[p]
	\small
	\centering
	\includegraphics[width=1.0\textwidth]{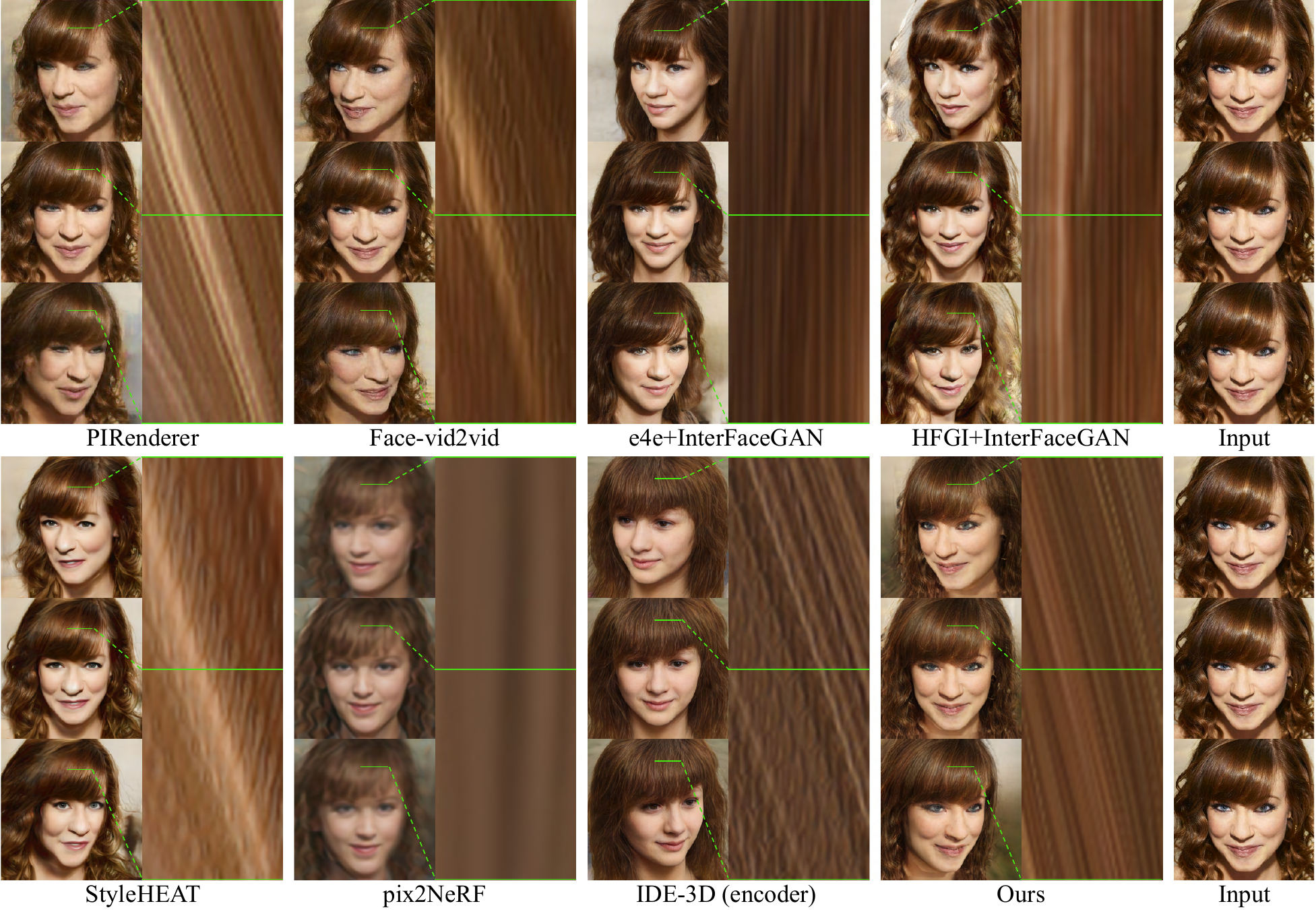}
	\caption{More pose editing comparisons. \textbf{Best viewed with zoom-in and see the animations.}}
	\label{fig:compare_more1}
\end{figure*}

\begin{figure*}[p]
	\small
	\centering
	\includegraphics[width=1.0\textwidth]{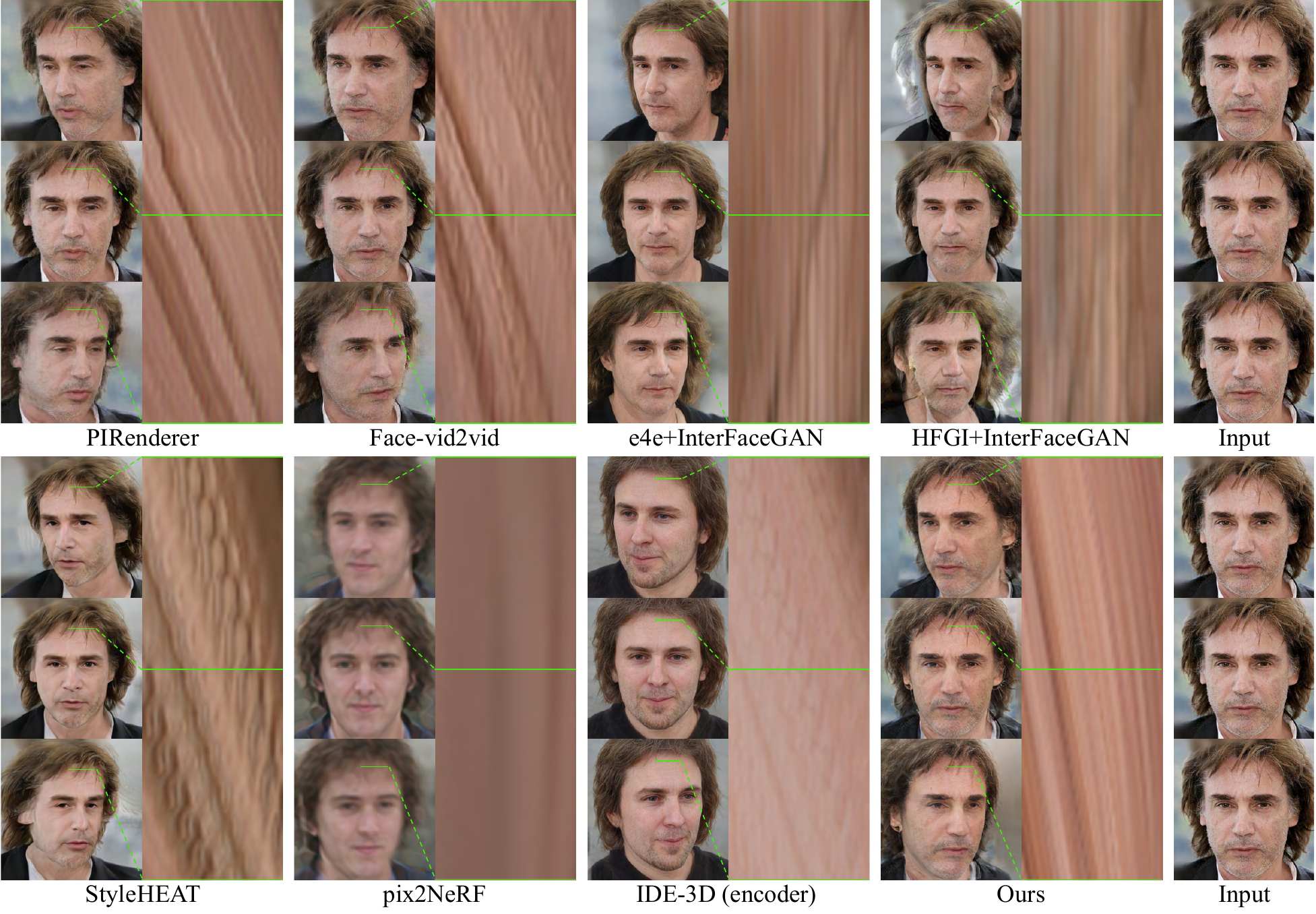}
	\caption{More pose editing comparisons. \textbf{Best viewed with zoom-in and see the animations.}}
	\label{fig:compare_more2}
\end{figure*}

Figure~\ref{fig:compare_more1} and \ref{fig:compare_more2} show more comparisons between our method and the previous methods. Our method can well preserve fine details in the original images and produces their novel views with more strict 3D consistency compared to the others. \textbf{Please see the \href{https://yudeng.github.io/GRAMInverter}{project page} for animations.}

We further compare with the full inversion pipeline of IDE-3D, which adopts the EG3D structure and leverages optimization-based inversion (\ie encoder-based initialization + pivot tuning~\cite{roich2021pivotal}). The results and a visual example are shown in Tab.~\ref{tab:compare_ide3d} and Fig.~\ref{fig:compare_ide3d}. The PSNR, LPIPS, and ID$_{nv}$ are calculated on the first $100$ instances in the CelebA-HQ, and the PSNR$_{mv}$ on $50$ instances. Our method performs slightly worse than the state-of-the-art optimization-based method on image reconstruction quality, but shows better novel view results and 3D consistency, and has dramatically faster inference speed.

\subsection{More Applications}

\begin{figure*}[p]
	\small
	\centering
	\includegraphics[width=1.0\textwidth]{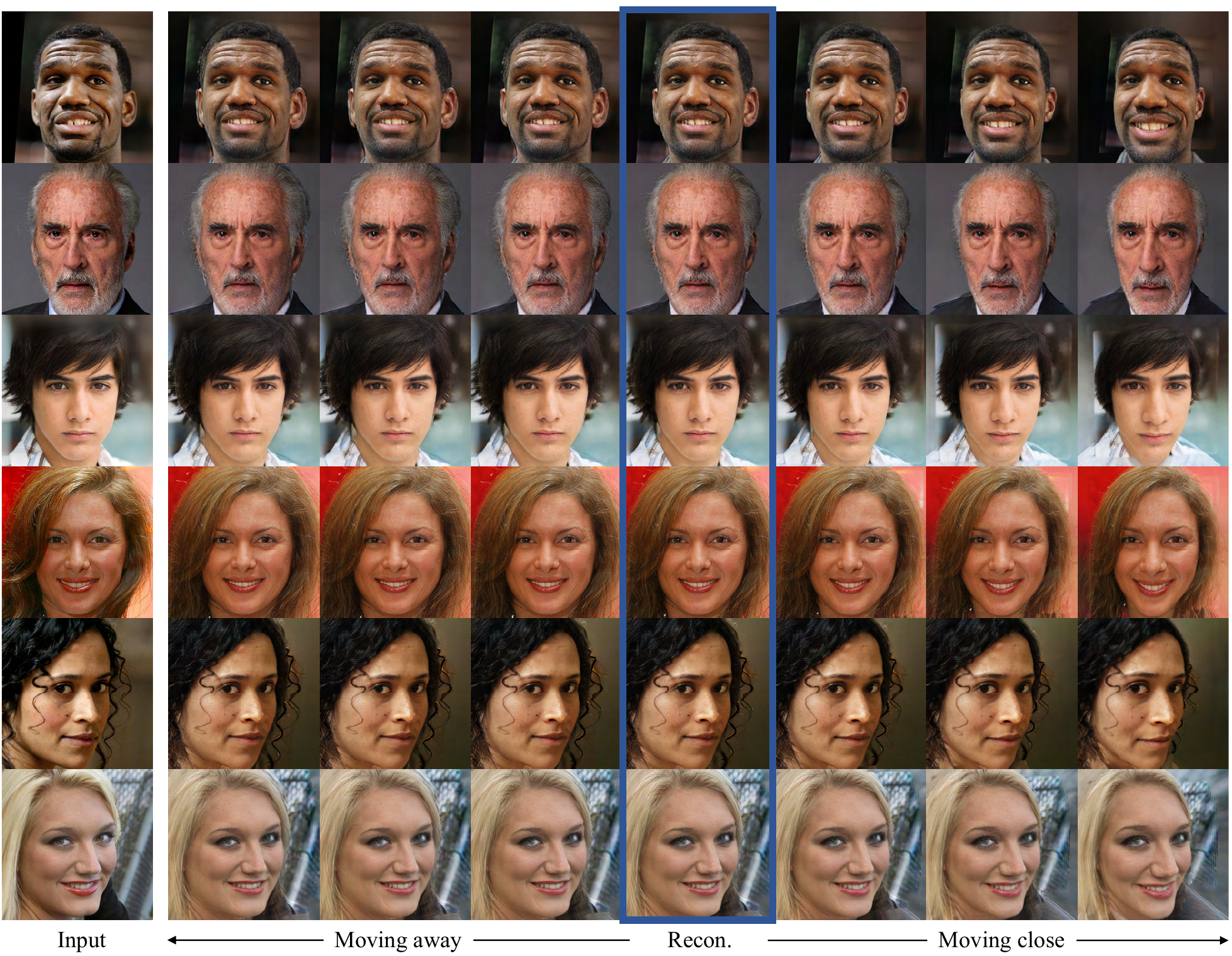}
	\caption{Dolly zoom effect of the given portraits produced by our method. \textbf{See the \href{https://yudeng.github.io/GRAMInverter}{project page} for animations.}}
	\label{fig:zoom}
\end{figure*}

\paragraph{Dolly zoom effect.} Since our method is based on GRAM~\cite{deng2022gram} that leverages the radiance manifolds representation, we can explicitly move the camera towards or away from a subject, and adjust the camera fov accordingly to ensure that the size of a portrait in the synthesized image stays a constant. In this way, we can generate a sequence of images under different levels of camera distortions, which is known as the dolly zoom effect\footnote{\href{https://en.wikipedia.org/wiki/Dolly\_zoom}{https://en.wikipedia.org/wiki/Dolly\_zoom}}. It can hardly be achieved by 2D-GAN based face editing methods without explicit camera modeling. Examples of this effect generated by our method are shown in Fig.~\ref{fig:zoom}. \textbf{Animations can be found in the \href{https://yudeng.github.io/GRAMInverter/}{project page}.}

 \begin{figure}[t]
	\small
	\centering
	\includegraphics[width=1.0\columnwidth]{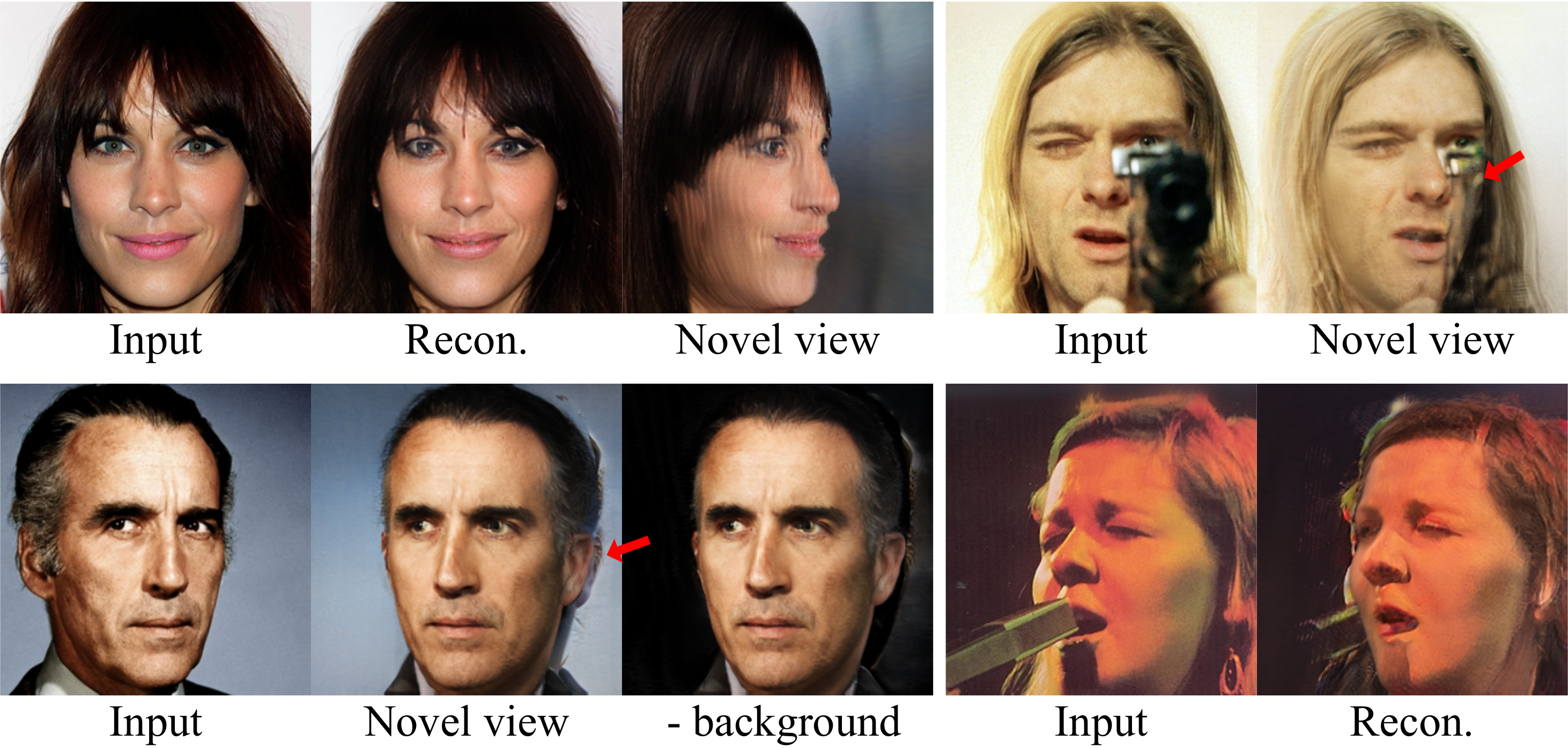}
	\caption{Limitations of our method. It can produce layered artifacts under large viewing angles. We also observe a ghosting artifact for certain subjects where the background contains the appearance of ears. In addition, it cannot well handle occlusions and out-of-distribution data. }
	\label{fig:artifact}
\end{figure}  

 \begin{figure}[t]
	\small
	\centering
	\includegraphics[width=1.0\columnwidth]{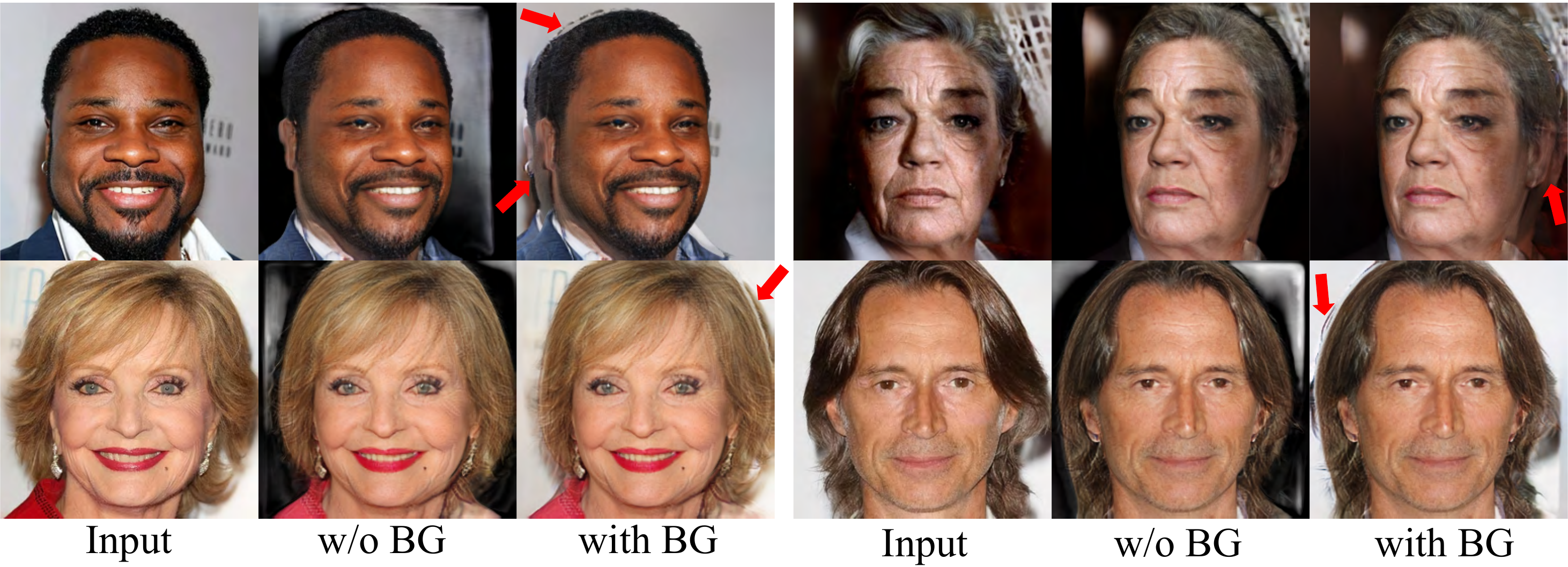}
	\caption{More comparisons between rendering with or without the background plane. \textbf{Best viewed with zoom-in.}}
	\label{fig:bg}
\end{figure}

\begin{figure*}[p]
	\small
	\centering
	\includegraphics[width=1.0\textwidth]{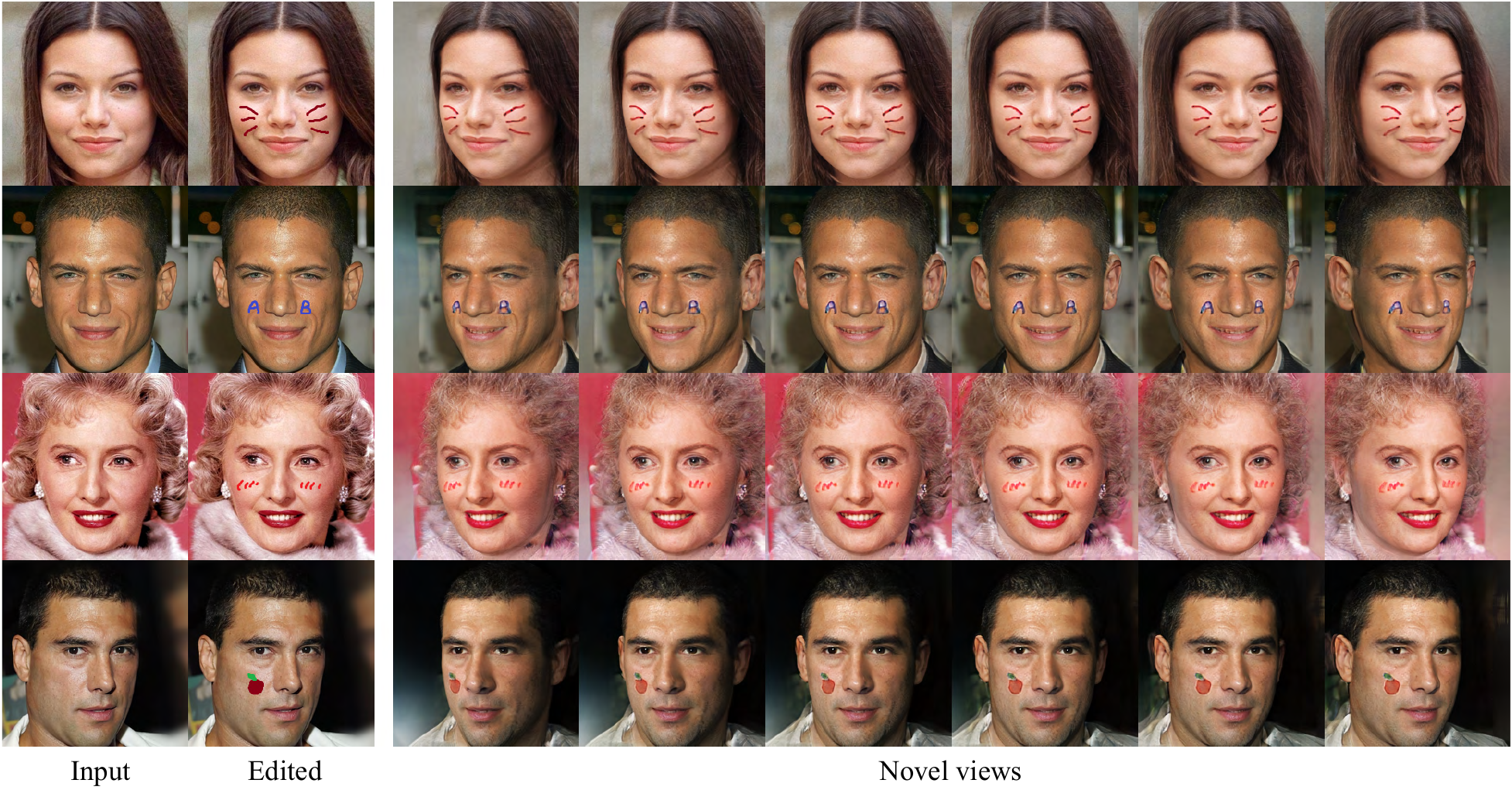}
	\caption{3D-consistent portrait editing results by our method. \textbf{See the \href{https://yudeng.github.io/GRAMInverter}{project page} for animations.}}
	\label{fig:edit}
\end{figure*}

\paragraph{3D-consistent editing.} Our method can also be applied to 3D-consistent interactive portrait editing thanks to its ability to preserve fine image details. Specifically, given a real portrait image, we can draw some arbitrary patterns on it and send the edited result to our GRAMinverter for reconstruction and novel view synthesis. As shown in Fig.~\ref{fig:edit}, our method can well preserve the drawn patterns on the input images and generate their 3D-consistent novel views bearing these patterns. \textbf{Corresponding animations are in the \href{https://yudeng.github.io/GRAMInverter/}{project page}.}

\section{Limitations and Future Works}
We thoroughly discuss the limitations of our method and possible future solutions to improve it. 

Our method adopts the radiance manifolds representation. Although it helps us to synthesize novel views with strong 3D consistency, it can produce layered artifacts at large viewing angles as shown in Fig.~\ref{fig:artifact}. This artifact could be alleviated to some extent by using more profile images as training data. In addition, it could also be reduced by leveraging alternative 3D representations, such as some recently proposed efficient NeRF representations~\cite{muller2022instant,skorokhodov2022epigraf,hong2022eva3d}. However, it is still unclear how to effectively incorporate these representations for high-quality and efficient novel view synthesis of monocular portraits. 

We also observed ghosting artifacts in some cases where the background contains the appearance of ears. The major cause is that the background plane and the foreground subject share the same tri-plane generator so they might have similar appearance patterns in some regions. Some floating points can also be observed around the silhouette, which are mainly due to the wrong parallax provided by inaccurate coarse depth (geometry) estimated from the general inversion stage. These problems can be alleviated by only rendering the foreground subject as shown in Fig.~\ref{fig:artifact}, or using an extra image generator to synthesize the background. We show more comparisons with or without the background in Fig.~\ref{fig:bg}. Clearly, removing the background largely reduces the layered artifacts and the floating points.

Besides, our method cannot well handle occlusions and tends to interpret them as textures clinging to the face as shown in Fig.~\ref{fig:artifact}. One possible solution is to leverage an extra face segmentation network to mask out the occluded regions and let the model only focus on reconstructing the portrait region. Our method can also produce inferior results for out-of-distribution input with large poses and abnormal lighting. The synthesized images may also have a global color shift compared to the input in certain cases. We believe these problems can be mitigated by training on larger-scale datasets with carefully tuned loss weights. In addition, our method cannot well handle complex lighting effect when varying the camera pose, such as specular reflectance. More dedicated 3D representations~\cite{guo2022nerfren} are required to tackle this problem.

Finally, our method does not support editing of attributes like expression, due to the learned details being aligned with the original input image. This problem can be tackled by introducing a distortion-aware detail reconstructor similarly as done by some recent 2D GAN inversion methods~\cite{wang2022high}, or leveraging a 3D representation that handles dynamic changes~\cite{park2021nerfies,wu2022anifacegan}. We leave these explorations as future works.

\section{Ethics Consideration}
The goal of this paper is efficient large-scale virtual avatar creation. It does not intend to create misleading or deceptive content. However, it could still be potentially misused for impersonating humans. In particular, the 3D-consistent synthesized portraits might be used to fool the 3D face recognition system that relies on multiview consistency. We condemn any behavior to create such harmful content. Currently, the synthesized portraits by our method contain certain visual artifacts that can be identified by humans and some deepfake detection methods. We encourage to apply this method for learning more advanced forgery detection approaches to avoid potential misusage.

\end{document}